\documentclass{article}

\PassOptionsToPackage{numbers}{natbib}
\PassOptionsToPackage{table}{xcolor}
\usepackage[main,final]{neurips_2026}

\usepackage[utf8]{inputenc}
\usepackage[T1]{fontenc}
\usepackage{url}
\usepackage{booktabs}
\usepackage{amsfonts}
\usepackage{amsmath}
\usepackage{mathtools}
\usepackage{nicefrac}
\usepackage{microtype}
\usepackage{xcolor}
\usepackage{multirow}
\usepackage{adjustbox}
\usepackage{graphicx}
\usepackage{placeins}
\usepackage{float}
\usepackage{arydshln}
\usepackage{subcaption}
\usepackage{wrapfig}
\usepackage{algorithm}
\usepackage{algorithmic}
\usepackage[most]{tcolorbox}
\usepackage{tikz}
\usepackage[colorlinks=true,citecolor=teal]{hyperref}
\usepackage[capitalize]{cleveref}

\crefname{section}{Sec.}{Secs.}
\Crefname{section}{Section}{Sections}
\Crefname{table}{Table}{Tables}
\crefname{table}{Tab.}{Tabs.}
\crefname{figure}{Fig.}{Figs.}

\def\ie{\textit{i.e.}}
\def\eg{\textit{e.g.}}

\definecolor{DarkGreen}{rgb}{0.0,0.5,0.0}
\definecolor{DarkYellow}{RGB}{180,140,40}
\definecolor{rxklightblue}{rgb}{0.92,0.95,1.0}
\definecolor{rxkdarkdarkblue}{rgb}{0.05,0.2,0.5}

\newcommand{\mathboxgreen}[1]{\colorbox{green!10}{$\displaystyle #1$}}
\newcommand{\mathboxyellow}[1]{\colorbox{yellow!20}{$\displaystyle #1$}}

\newenvironment{prompt_gray}[1][]
  {\begin{tcolorbox}[
      enhanced,
      breakable,
      boxrule=0.5pt,
      arc=4pt,
      left=2pt,
      right=2pt,
      bottom=2pt,
      top=2pt,
      rounded corners
    ]
    \textbf{#1.}
    \small\itshape}
  {\end{tcolorbox}}

\title{Universal Cross-Prompt Adversarial Attacks on Promptable Concept Segmentation}

\author{
 Ziqi Zhou$^{1}$, Yifan Hu$^{2}$, Yufei Song$^{2}$, Haowen Jiang$^{3}$, Xianlong Wang$^{4}$, \\ \textbf{Shengshan Hu}$^{2}$, \textbf{Dezhong Yao}$^{3}$, \textbf{Leo Yu Zhang}$^{5}$
 \\
 $^{1}$ College of Computer Science, Chongqing University \\
 $^{2}$ School of Cyber Science and Engineering, Huazhong University of Science and Technology \\
 $^{3}$ School of Computer Science and Technology, Huazhong University of Science and Technology\\
 $^{4}$ Department of Computer Science, City University of Hong Kong \\
 $^{5}$ School of Information
and Communication Technology, Griffith University\\
\footnotesize{\texttt{zhouziqi@cqu.edu.cn}},
\footnotesize{\texttt{\{hyf1009,yufei17,jianghaowen,hushengshan,dyao\}@hust.edu.cn}}
\\
\footnotesize{\texttt{xianlong.wang@my.cityu.edu.hk}},
\footnotesize{\texttt{leo.zhang@griffith.edu.au}}
}

\let\circleone\filledcircle
\let\circletwo\filledcircle
\let\circlethree\filledcircle

\begin{document}

\maketitle

\vspace{-0.6cm}
\begin{figure}[H]
  \centering
  \includegraphics[scale=0.62]{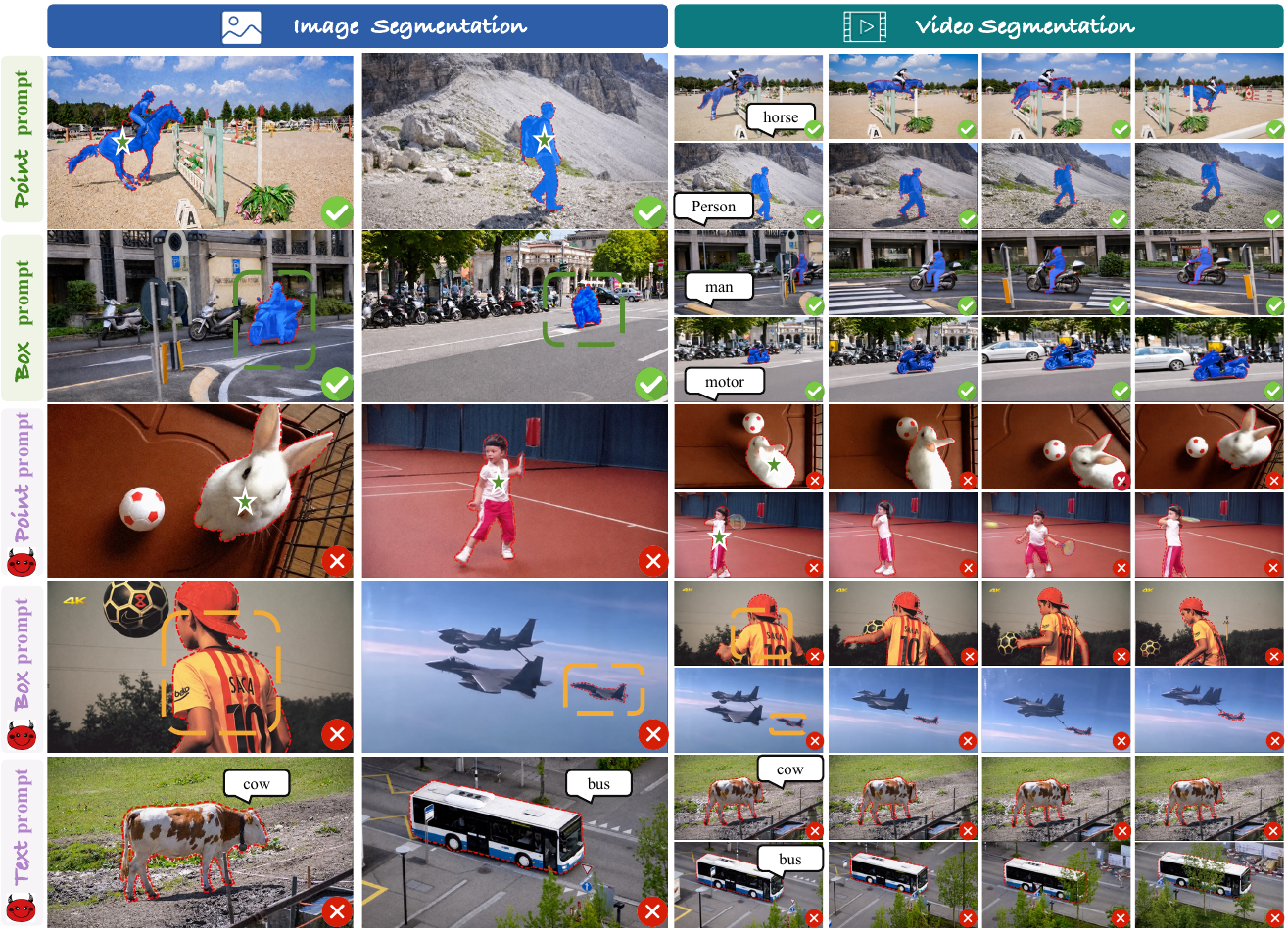}
  \caption{Visualization of SAM3 segmentation results on benign samples and adversarial examles generated by our method under point, box, and text prompts. The first and second rows correspond to benign samples, while the third to fifth rows correspond to adversarial examles. For better visualization, we use red dashed contours to indicate the ground-truth masks of the target objects, and blue highlights to indicate the regions segmented by SAM3.
}
  \label{fig:visualization}
  \vspace{-0.4cm}
\end{figure}

\begin{abstract}
The \textit{\textbf{S}egment \textbf{A}nything \textbf{M}odel} (SAM) achieves remarkable performance in visual segmentation. 
The latest SAM3 extends promptable segmentation to concept-level prediction, broadening the scope of segmentation foundation models.
While recent works reveal that SAM and SAM2 are vulnerable to adversarial examples, the robustness of SAM3 under the concept segmentation paradigm remains  unexplored. 
In addition, existing adversarial attacks on SAM-series models exhibit limited cross-prompt transferability. 
To this end, 
we propose AdvPCS, a universal cross-prompt adversarial attack for \textit{\textbf{P}romptable \textbf{C}oncept \textbf{S}egmentation} (PCS), 
including a min-max prompt optimization strategy, a global–local perception deception attack, and a temporal transition deviation attack. 
Specifically, we first identify the hardest-to-attack prompts via the min-max bilevel optimization. 
In the inner maximization, we enhance diversity over candidate point, box, and text prompts. 
In the outer minimization, we select prompts with the highest responses based on the confidence scores output by the detector.
Given the selected prompts, we apply the perception deception attack to minimize both global and local existence probabilities under joint prompting, and employ the temporal memory misalignment attack to maximize inter-frame semantic inconsistency and corrupt memory pointers. 
Extensive experiments on four benchmark datasets show that a single \textit{universal adversarial perturbation} (UAP) generated by our method generalizes across frames from different videos and achieves strong attack performance under point, box, and text prompts. 
In particular, it reduces the average mIoU of various PCS models on the SA-CO dataset to below $5\%$ under text prompts, demonstrating strong attack ability.
The code is available at \url{https://github.com/alphanull-cqu/AdvPCS}.
\end{abstract}

\section{Introduction}\label{sec:inroduction}

The Segment Anything Model (SAM) and its variants~\cite{carion2025sam,kirillov2023segment,ravi2024sam2,zhang2026efficient} have demonstrated strong generalization in visual segmentation.
SAM~\cite{kirillov2023segment} introduces \textit{promptable visual segmentation} (PVS) paradigm, enabling general-purpose image segmentation via prompts.
Moreover, 
SAM2~\cite{ravi2024sam2} extends this framework to videos by incorporating memory mechanisms that ensure temporally consistent segmentation and tracking. 
To achieve \textit{promptable concept segmentation} (PCS), 
the recent SAM3~\cite{carion2025sam} introduces a unified framework for detection, tracking, and segmentation, which takes simple noun phrases as input to understand, track, and precisely segment target objects. 
Compared with SAM and SAM2, SAM3 not only models texture and boundary cues but also identifies all instances in the scene that match the given concept, outputs their segmentation masks with unique identities.

SAMs are shown to be vulnerable to adversarial examples~\cite{long2025robust,lu2024unsegment,song2025seg,xia2024transferable,zhou2023advclip,zhou2023downstream,zhou2025numbod, zhou2024securely}, where attackers add carefully crafted and imperceptible perturbations to target frames to induce incorrect predictions. 
Existing attacks on SAM mainly focus on PVS. 
They either corrupt the image encoder features without relying on prompts~\cite{croce2024segment,zheng2024black} or optimize adversarial perturbations with multiple random prompts~\cite{zhou2024darksam} to achieve cross-prompt transferability. 
A recent work, UAP-SAM2~\cite{zhou2025sam2} disrupts temporal semantic consistency, significantly degrading video segmentation performance in SAM2. 
Although these methods perform well on SAM and SAM2 under a specific prompt type, their effectiveness drops substantially under diverse prompt types (See \cref{tab:comparison_video_image}). 
Moreover, to the best of our knowledge, the robustness of emerging PCS models remains  unexplored.

Different from SAM~\cite{kirillov2023segment} and SAM2~\cite{ravi2024sam2}, which rely on user-provided pixel-level prompts (\eg, points or boxes), 
SAM3~\cite{carion2025sam} introduces a detector to determine whether a concept-level object exists in the current frame and then segments all regions associated with the queried concept.
To study this new property, we conduct exploratory experiments in \cref{sec:insights} from the perspectives of improving both attack effectiveness and transferability. These experiments reveal two key insights: (1) deceiving the perception mechanism in SAM3 improves attack effectiveness, and (2) mining the hardest-to-attack prompts for attack optimization enhances cross-prompt transferability.
Moreover, for video tasks, effectively disrupting semantic consistency across consecutive frames remains a challenge issue.

In this paper, we propose AdvPCS, a brand-new cross-prompt universal adversarial attack for promptable concept segmentation models. 
AdvPCS integrates a min-max bilevel adversarial prompt optimization strategy, a global–local perception deception attack, and a temporal transition deviation attack.
Our method targets both cross-prompt transferability and perception-level distortion, aiming to generate a   \textit{universal adversarial perturbation} (UAP) that generalizes across diverse prompts and frames.
To enhance cross-prompt transferability, we design a min–max bilevel adversarial prompt optimization strategy that identifies the hardest-to-attack prompts for optimization. 
In the inner maximization, we increase diversity over candidate point, box, and text prompts. 
In the outer minimization, we select the three types of prompts with the highest responses based on the probability scores from the detector and jointly optimize them in subsequent steps.
For perception-level distortion, we design a global–local perception deception attack that jointly minimizes the global and local presence probabilities under the three  types of prompts, while constraining their pairwise differences to prevent the perturbation from overfitting to any single prompt. 
We then design a temporal transition deviation attack that disrupts semantic consistency across consecutive frames and deliberately misguides memory pointers, thereby enhancing cross-frame attack effectiveness.

To better evaluate the transferability, we train UAPs only on the SA-CO~\cite{carion2025sam} dataset and evaluate them on other datasets under point, box, and text prompts. 
\cref{fig:visualization} shows visualizations of the adversarial examples generated by our method. 
The qualitative results in the last three columns show that these samples cannot be effectively segmented by SAM3 under all three types of prompts.
The quantitative results in \cref{sec:experiments} demonstrate that our method achieves strong attack performance on both video and image segmentation tasks across four datasets.
Our main contributions are summarized as follows:
\begin{itemize}
\item 
We propose AdvPCS, a novel cross-prompt universal adversarial attack for promptable concept segmentation models. 
We generate a single UAP only on the SA-CO dataset, which generalizes across different videos, frames, prompt types, and prompt instances, and achieves strong attack performance against PCS models.

\item 
We design a min–max bilevel adversarial prompt optimization strategy that identifies the most robust prompt instances across text, point, and box modalities, thereby significantly improving cross-prompt transferability of the UAP.

\item 
Extensive experiments on four datasets show that our method transfers across frames from different videos and achieves strong attack performance under point, box, and text prompts.
Comparative and defense experiments further validate its effectiveness and superiority.

\end{itemize}

\vspace{-0.4cm}
\section{Related Works}
\subsection{Segment Anything Models}
Segment Anything Model (SAM)~\cite{kirillov2023segment} first formulates image segmentation as an interactive prompt learning problem and enables general-purpose segmentation with prompts such as points and boxes. 
SAM2~\cite{ravi2024sam2} extends this framework to videos by introducing Transformer-based~\cite{vaswani2017attention} temporal modeling and a memory mechanism, enabling temporally consistent segmentation and tracking across frames. 
It unifies image and video processing within a single architecture.
SAM3~\cite{carion2025sam} introduces \textit{Promptable Concept Segmentation} (PCS) with a decoupled detection and tracking architecture, where both components share a unified perception encoder. 
The detector performs open-vocabulary object detection, while the tracker inherits the memory mechanism of SAM2 for temporal identity tracking. The perception encoder processes each image once and produces prompt-independent features shared by both modules.
Benefit from their strong segmentation capabilities~\cite{chen2023sam,ke2023segment,kweon2024sam,wang2026x,zhang2026efficient}, they are widely adopted in applications such as medical image analysis~\cite{cheng2023sam,sengupta2025sam} 
and autonomous driving~\cite{camarena2025ad,cheng2025ur}.

\subsection{Adversarial Attacks on SAMs}
Recent works~\cite{han2023sam,jiang2024cross,long2025robust,lu2024unsegment,qiao2023robustness,shen2024practical,xia2024transferable} study adversarial examples to mislead SAM and its downstream tasks.
AttackSAM~\cite{zhang2023attack} exploits SAM’s mask generation mechanism to craft targeted perturbations that mislead object segmentation.
Although these methods effectively mislead SAM under fixed prompt settings, their effectiveness drops significantly as prompts vary. 
Consequently, recent studies~\cite{croce2024segment,huang2024segment,liu2024cross,zhou2025sam2,zhou2024darksam} shift focus to cross-prompt attacks against SAMs. 
Zheng et al.~\cite{zheng2024black} and Croce et al.~\cite{croce2024segment} propose transferable attacks that operate solely on the image encoder, removing dependence on specific prompts.
DarkSAM~\cite{zhou2024darksam} further introduces a hybrid frequency–spatial adversarial attack to enhance cross-prompt attack effectiveness.
Recently, UAP-SAM2~\cite{zhou2025sam2} shows that attacks designed for SAM1 fail on SAM2 due to semantic consistency across consecutive frames.
In this work, we focus on the robustness of emerging PCS models and aim to generate a single UAP that effectively fools these models across different frames, prompt types, and prompt instances.

\vspace{-0.4cm}
\section{Methodology}\label{sec:methodology}

\subsection{Problem Formulation}\label{sec:problem}
Given an input sequence of frames $\mathbf{X} = \{x_i\}_{i=1}^N$ with $N$ frames and prompts  $ \mathbb{P} = \{p_i^{j}\}_{i=1}^N$, $j\in\{\mathrm{text},\mathrm{point},\mathrm{box}\}$, where $j \in {t, p, b}$ denotes the prompt type, corresponding to text, point, and box, respectively.
Let $f_{\theta}$ denotes SAM3 or its variants.
$f_{\theta}$ predicts segmentation masks $\mathbf{Y} = \{y_i\}_{i=1}^N$ for each frame $x_i$. 
For a frame $x_i$, a pixel at coordinates $(m,n)$, denoted as $x_i^{mn}$, is considered part of the masked region if its corresponding mask value $y_i^{mn}$ exceeds zero.
SAM3 consists of the following modules: an image encoder that extracts visual features from each input frame, a detector that predicts object presence based on the extracted features, a memory fusion module that integrates current-frame features with historical memory representations, and a tracker that generates candidate masks and selects the optimal mask as the final segmentation result.
Following~\cite{zhou2025sam2}, we assume that the attacker has access to the open-source SAM3 and can leverage public datasets to generate adversarial perturbations.
The attacker aims to craft a UAP $\delta$ for each video frame such that the target object cannot be effectively segmented by SAM3. Meanwhile, $\delta$ remains imperceptible by enforcing an $l_p$-norm constraint bounded by a predefined perturbation budget $\epsilon$.

\subsection{Motivation}\label{sec:insights}
\begin{figure*}[!h]
    \centering
    \begin{subfigure}[t]{0.43\textwidth}
        \centering
        \includegraphics[width=\textwidth]{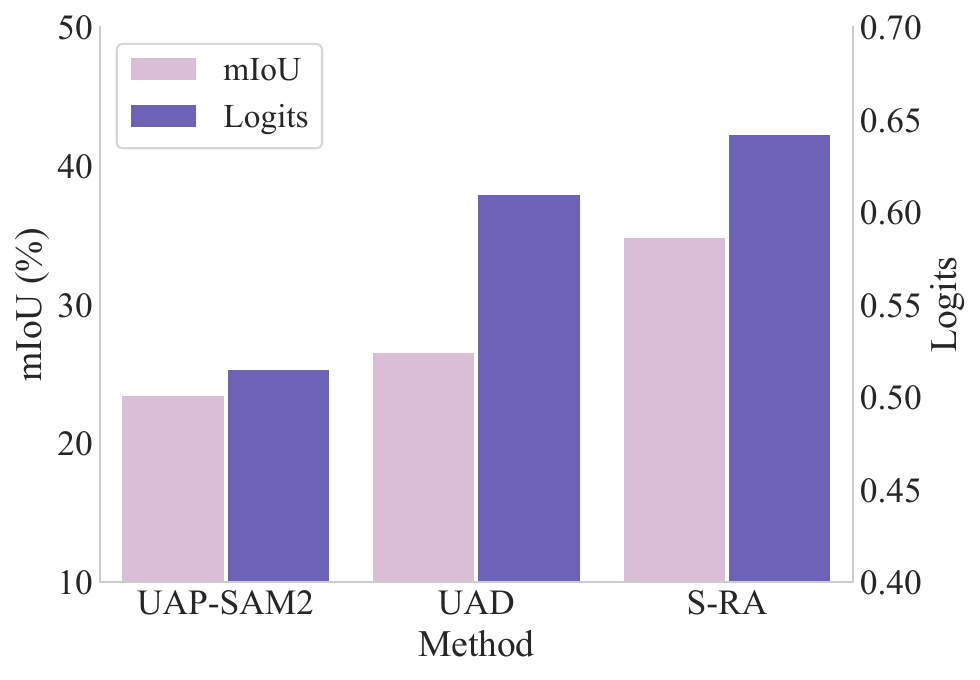}
        \caption{Logit and mIoU comparison}
        \label{fig:logit}
    \end{subfigure}
    \hfill
    \begin{subfigure}[t]{0.54\textwidth}
        \centering
        \includegraphics[width=\textwidth]{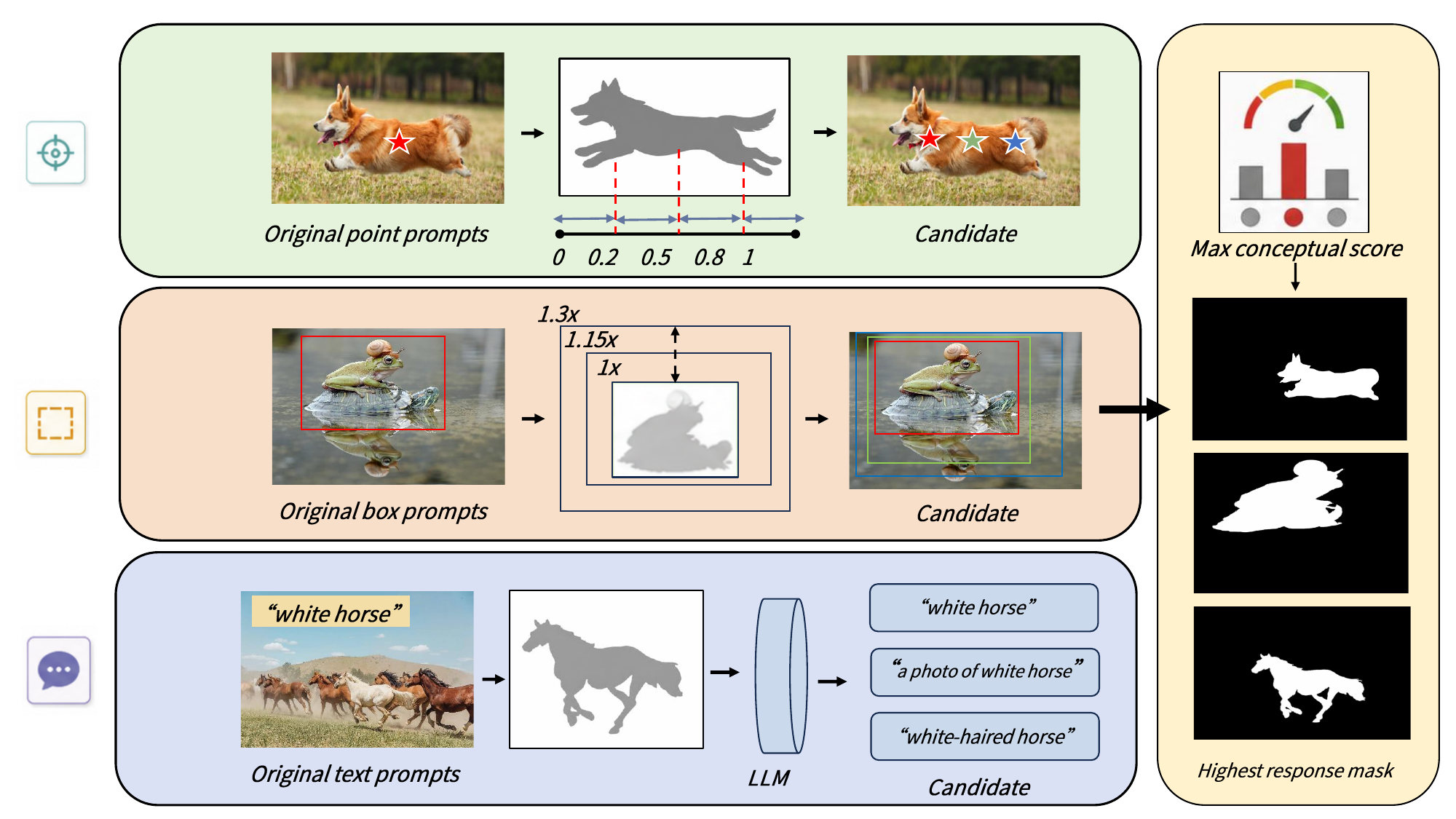}
        \caption{Hardest-to-attack prompts mining}
        \label{fig:minmax}
    \end{subfigure}
    \caption{Illustration of the key insights for designing effective attacks against PCS}
    \label{fig:insight}
    \vspace{-0.4cm}
\end{figure*}

Motivated by the structural characteristics of SAM3,
we design the attack from two perspectives: improving its effectiveness and transferability.
Here are two key insights behind our method:

\noindent\textbf{Insight I: Deceiving the perception mechanism improves attack effectiveness.}
Compared to previous models, SAM3 introduces a detector that jointly recognizes and localizes objects associated with open-vocabulary prompts, providing stronger semantic grounding. However, existing attacks do not explicitly target concept-level perception induced by textual prompts. Under open-vocabulary settings, ambiguities such as polysemy and subjective descriptions make it difficult for conventional feature-level perturbations~\cite{croce2024segment,zheng2024black} or global semantic attacks~\cite{madry2017towards} to consistently disrupt all candidate objects.
We evaluate three state-of-the-art (SOTA) attacks for SAM and SAM2, including UAP-SAM2~\cite{zhou2025sam2}, UAD~\cite{lu2024unsegment}, and S-RA~\cite{shen2024practical}, by generating adversarial examples on the SA-CO dataset to attack SAM3. 
As shown in \cref{fig:logit}, the average results across the three types of prompts (\ie, point, box, and text) reveal a clear correlation between the degradation in segmentation performance and the decrease in the detector’s existence confidence. This observation motivates us to design attacks that directly target the perception mechanism to enhance attack effectiveness.

\vspace{-0.2cm}

\noindent\textbf{Insight II: Hardest-to-attack prompt mining improves cross-prompt transferability.}
In addition to input variations, diverse prompt types and instances pose significant challenges for transferability. Existing attacks typically optimize for a single prompt type, and their effectiveness degrades substantially when the prompt type changes. Similarly, feature-space attacks become less effective when prompt information is rich.
Inspired by adversarial training~\cite{madry2017towards}, we improve transferability from two perspectives: (i) jointly optimizing across point, box, and text prompts to enhance robustness under heterogeneous prompts; and (ii) identifying the hardest-to-attack prompt within each type to further strengthen transferability.
As illustrated in \cref{fig:minmax}, we design three mining strategies for different prompt types.
Specifically,
\circleone{1} for \textit{text prompts}, we expand a given prompt into multiple semantically similar candidates using LLMs (\eg, GPT5.4) and select the one that yields the highest object existence score;
\circletwo{2} for \textit{point prompts}, we construct multiple candidates from the foreground mask, including left-biased, center, and right-biased points based on spatial quantiles, and select the point that maximizes the response;
\circlethree{3} for \textit{box prompts}, we generate a default bounding box from the foreground mask and further create scaled variants (\eg, enlarged by $15\%$ and $30\%$) while keeping the center fixed.
The above mining strategy identifies the hardest-to-attack prompts for each type, which are then used for subsequent attack optimization.

\begin{prompt_gray}[Takeaway]
An effective attack on SAM3 requires deceiving its global–local perception mechanism while achieving strong transferability across videos, frames, prompt types, and prompt instances.
\end{prompt_gray}

\subsection{AdvPCS: A Complete Illustration}
\begin{figure*}[!t]

    \centering
    \includegraphics[scale=0.42]{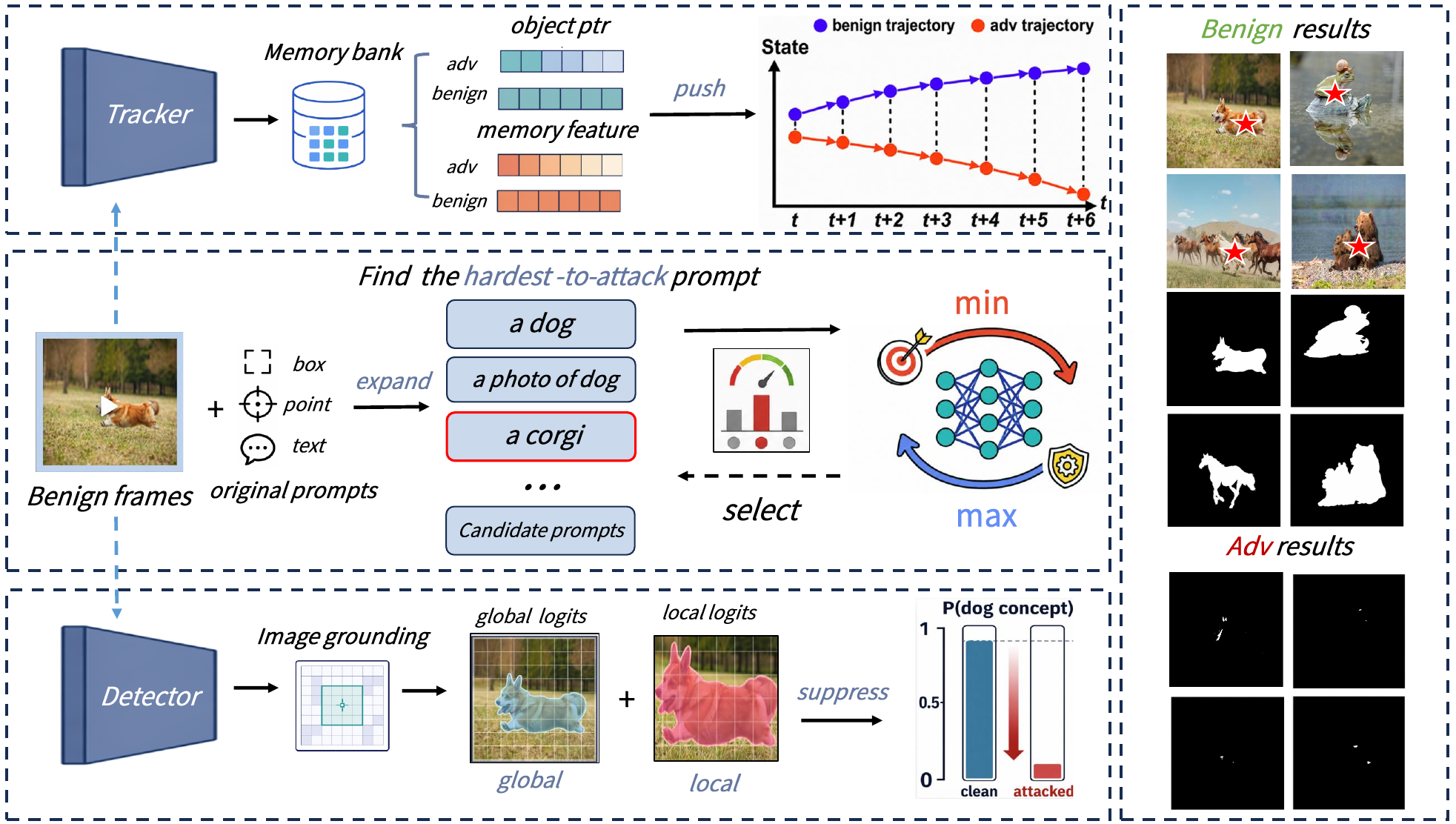}
    \caption{The framework of AdvPCS
    }
    \label{fig:pipeline}
    \vspace{-0.4cm}
\end{figure*}

In this section, we propose AdvPCS, a brand-new cross-prompt universal adversarial attack for PCS models. 
The pipeline of AdvPCS illustrated in \cref{fig:pipeline} integrates a min–max bilevel adversarial prompt optimization strategy, a perception-deception attack, and a temporal transition deviation attack.
The overall optimization objective of AdvPCS is as follow: 
\begin{equation}
\mathcal{J}_{total}= \mathcal{J}_{pa} + \mathcal{J}_{ta},
\label{eq:jt}
\end{equation}
where $\mathcal{J}_{pa}$ is the global–local perception deception attack loss and $\mathcal{J}_{ta}$ denotes the temporal transition deviation attack loss.

\noindent\textbf{Min--max bilevel adversarial prompt optimization.}
To improve cross-prompt transferability, we propose a min--max bilevel adversarial prompt optimization strategy that identifies the hardest-to-attack prompts for optimization.
In the inner maximization, we increase diversity over candidate point, box, and text prompts. 
In the outer minimization, we select the highest-response prompts from the three types based on the probability scores produced by the perception encoder, and jointly optimize them to further enhance transferability across both prompt types and prompt instances.
This optimization process can be formulated as:
\begin{equation}
\min_{\|\delta\|_p \le \epsilon} 
\max_{p_i^j \in \mathbb{P}}
\sum_{i=1}^{N} \sum_{j \in \{t,p,b\}}
\mathcal{L}\big(f_{\theta}(x_i + \delta,\; p_i^j),\; y_i\big),
\end{equation}
where 
$\mathcal{L}(\cdot)$ is the segmentation loss. 

\noindent\textbf{Global–local perception deception attack.}
We aim to degrade SAM3’s segmentation capability by deceiving its perception stage. From an analysis of its perception module, SAM3 employs a detector that produces a global perception logit and local perception logits to recognize and localize the target object specified by the input prompt. Specifically, the global perception logit captures the overall confidence of the target’s presence in a frame, while the local perception logits are more sensitive to prompt-conditioned local features.
Accordingly, we disrupt perception by suppressing both global and local confidence of the target across all the three prompt types. We further impose a consistency degradation constraint on prompt-dependent local responses to prevent the perturbation from overfitting to any single prompt.
Hence, we define the perception-deception objective as follows:
\begin{equation}
\mathcal{J}_{pa} = \mathcal{J}_{global} +  \mathcal{J}_{local} + \mathcal{J}_{reg},
\end{equation}
where $\mathcal{J}_{global}$ denotes the global existence weakening loss, $\mathcal{J}_{local}$ denotes the local existence weakening loss, and $\mathcal{J}_{reg}$ denotes the cross-prompt consistency regularization loss.

Given an input frame $x_i$, we add a UAP $\delta$ to obtain the adversarial frame $\tilde{x}_i = x_i + \delta$. Under text, point, and box prompts, the detector produces the corresponding global perception logits $r_i^t$, $r_i^p$, and $r_i^b$, 
as well as local perception logits $z_i^t$, $z_i^p$, and $z_i^b$, where the superscripts $t$, $p$, and $b$ denote the text, point, and box prompt settings, respectively. 
We first suppress the global existence confidence of the target object under three types of prompts. The $\mathcal{J}_{global}$ is expressed as:
\begin{equation}
\mathcal{J}_{global}
=
\frac{1}{N}\sum_{i=1}^{N}
\left(
\left\| r_i^t - \tau  \right\|_2^2
+
\left\| r_i^p - \tau  \right\|_2^2
+
\left\| r_i^b - \tau  \right\|_2^2
\right),
\end{equation}
where $\tau <0$ is a predefined low target logit value. 
By uniformly suppressing global perception logits toward a shared low-confidence target, we weaken SAM3’s ability to recognize the target object globally.
We then further suppress the local perception logits under the three prompt types.
Our empirical results show that box prompts are more difficult to attack. Therefore, we place greater emphasis on box prompts during the optimization process.
The $\mathcal{J}_{local}$ is formulated as follows:
\begin{equation}
\mathcal{J}_{local}
=
\frac{1}{N}\sum_{i=1}^{N}
\left(
\left\| z_i^t - \tau  \right\|_2^2
+
\left\| z_i^p - \tau  \right\|_2^2
+
\lambda \left\| z_i^b - \tau  \right\|_2^2
\right),
\end{equation}
where $\lambda$ is a hyperparameter that controls the influence of box prompts during the adversarial optimization process.
To avoid overfitting to a single prompt type, we enforce consistent degradation of local responses across prompts in the probability space.
Let $p_i^t=\sigma(z_i^t)$, $p_i^p=\sigma(z_i^p)$, and $p_i^b=\sigma(z_i^b)$ denote the prediction probabilities under the text, point, and box prompts, respectively, where $\sigma(\cdot)$ is the sigmoid function.
We define the $\mathcal{J}_{reg}$ as follows:
\begin{equation}
\mathcal{J}_{reg}
=
\frac{1}{3N}\sum_{i=1}^{N}
\left(
\left\| p_i^t-p_i^p \right\|_2^2
+
\left\| p_i^t-p_i^b \right\|_2^2
+
\left\| p_i^p-p_i^b \right\|_2^2
\right).
\end{equation}

\noindent\textbf{Temporal transition deviation attack.}
Motivated by~\cite{zhou2025sam2}, which shows that memory weakens adversarial attacks in SAM2, we propose a temporal transition deviation attack that improves cross-frame transferability by disrupting temporal semantic consistency and misleading memory propagation.
Starting from the second frame, we feed both benign and adversarial frames into the tracker and extract the corresponding object pointers and memory features.
Let $\mathbf{o}_i$ denote the object pointer at the $i$-th frame, and let $\mathbf{m}_i$ denote the corresponding  features. 
For simplicity, we define the adjacent-frame pointer transition and memory transition as follows:
\begin{equation}
\Delta \mathbf{o}_i = \mathbf{o}_i - \mathbf{o}_{i-1}, \qquad
\Delta \mathbf{m}_i = \mathbf{m}_i - \mathbf{m}_{i-1}.
\end{equation}

Accordingly, we use $\Delta \mathbf{o}_i^{\mathrm{adv}}$ and $\Delta \mathbf{o}_i^{\mathrm{benign}}$ to denote the pointer transitions of the adversarial and benign sequences, respectively, and $\Delta \mathbf{m}_i^{\mathrm{adv}}$ and $\Delta \mathbf{m}_i^{\mathrm{benign}}$ are defined in the same way for the memory features. 
Unlike~\cite{zhou2025sam2}, which focuses on maximizing feature discrepancies between adjacent adversarial frames, we instead emphasize \textit{the deviation between adversarial and benign memory trajectories} at each time step. 
For each frame, we compute the temporal transition deviation $\pi_{i}$ between the adversarial and benign samples:
\begin{equation}
\pi_{i}
=
\underbrace{\mathboxgreen{\cos(\Delta \mathbf{o}_i^{\mathrm{adv}}, \Delta \mathbf{o}_i^{\mathrm{benign}})}}_{\textcolor{DarkGreen}{\textbf{Pointer Deviation}}}
+
\underbrace{\mathboxyellow{\cos(\Delta \mathbf{m}_i^{\mathrm{adv}}, \Delta \mathbf{m}_i^{\mathrm{benign}})}}_{\textcolor{DarkYellow}{\textbf{Feature Deviation}}}
\end{equation}
where $\cos(\cdot)$ denotes the cosine similarity loss. 
Finally, we obtain the objective of the temporal memory deviation attack $\mathcal{J}_{ta}$ as follows:

\begin{equation}
\mathcal{J}_{ta}
=
\frac{1}{N-1}\sum_{i=2}^{N} \pi_{i}
\end{equation}

By directly reducing the cosine similarity between the adversarial and benign temporal transitions, the adversarial sequence is progressively driven away from the benign temporal trajectory, so that the corrupted temporal cues accumulated from previous frames can further misguide subsequent predictions and amplify the attack effect over time.

\section{Experiments}\label{sec:experiments}

\subsection{Experimental Setup}\label{sec:experimental_setup}

\noindent\textbf{Datasets and models.} 
We evaluate the proposed attack on four public video segmentation benchmarks, including YouTube~\cite{xu2018youtube}, DAVIS~\cite{pont20172017}, MOSE~\cite{ding2023mose}, and SA-CO~\cite{carion2025sam}.
To further evaluate attack performance on image segmentation, following ADV-SAM2~\cite{zhou2025sam2}, we randomly sample 100 videos from YouTube~\cite{xu2018youtube}, DAVIS~\cite{pont20172017}, MOSE~\cite{ding2023mose}, and SA-CO~\cite{carion2025sam}, and extract 15 frames per video to construct image-level test sets. 
We resize all frames to $3 \times 1008 \times 1008$. We use official SAM3~\cite{carion2025sam}, SAM3.1~\cite{carion2025sam}, and EfficientSAM3~\cite{zeng2025efficientsam3} as victim models for evaluation.

\begin{table}[t]
\centering
\scriptsize
\setlength{\tabcolsep}{2.5pt}
\renewcommand{\arraystretch}{1.4}
\caption{The mIoU (\%) of AdvPCS under different settings.}
\label{tab:main_results}
\begin{adjustbox}{center,max width=\linewidth}
\begin{tabular}{c c || c c | c c | c c | c c || c c | c c | c c | c c}
\hline\hline
\rowcolor{gray!20}
\rule{0pt}{2.8ex} & &
\multicolumn{8}{c||}{Video Segmentation} &
\multicolumn{8}{c}{Image Segmentation} \\
\rowcolor{gray!20}
& &
\multicolumn{2}{c|}{SAM3} &
\multicolumn{2}{c|}{SAM3.1} &
\multicolumn{2}{c|}{E-SAM3} &
\multicolumn{2}{c||}{Avg} &
\multicolumn{2}{c|}{SAM3} &
\multicolumn{2}{c|}{SAM3.1} &
\multicolumn{2}{c|}{E-SAM3} &
\multicolumn{2}{c}{Avg} \\
\rowcolor{gray!20}
\multirow{-3}{*}{Prompt} &
\multirow{-3}{*}{Dataset} &
BoU & AoU & BoU & AoU & BoU & AoU & BoU & AoU &
BoU & AoU & BoU & AoU & BoU & AoU & BoU & AoU \\
\hline\hline
& YouTube
& 85.29 & 6.29 & 85.48 & 17.58 & 56.58 & 28.56 & 75.78 & 17.48
& 74.17 & 0.09 & 53.59 & 1.74 & 53.38 & 36.60 & 60.38 & 12.81 \\
\rowcolor{gray!10}
Point & MOSE
& 69.12 & 8.17 & 76.82 & 23.69 & 42.79 & 15.40 & 62.91 & 15.75
& 78.90 & 0.84 & 43.22 & 2.59 & 48.86 & 32.83 & 56.99 & 12.09 \\
& DAVIS
& 69.73 & 7.89 & 81.59 & 39.13 & 51.55 & 30.69 & 67.62 & 25.90
& 68.21 & 0.32 & 41.95 & 0.61 & 44.98 & 33.81 & 51.71 & 11.58 \\
\hdashline
\rowcolor{gray!10}
& YouTube
& 88.28 & 39.62 & 87.86 & 52.93 & 57.51 & 19.25 & 77.88 & 37.27
& 88.19 & 4.18 & 87.48 & 12.61 & 71.04 & 59.43 & 82.24 & 25.41 \\
Box & MOSE
& 75.15 & 25.20 & 82.73 & 43.20 & 39.74 & 13.37 & 65.87 & 27.26
& 86.54 & 7.95 & 79.15 & 19.08 & 64.48 & 54.80 & 76.72 & 27.28 \\
\rowcolor{gray!10}
& DAVIS
& 88.14 & 41.13 & 88.90 & 61.14 & 53.88 & 23.77 & 76.97 & 42.01
& 87.98 & 3.95 & 88.49 & 14.91 & 65.95 & 52.89 & 80.81 & 23.92 \\
\hdashline
Text & SA-CO
& 68.48 & 0.00 & 68.73 & 1.32 & 73.52 & 12.50 & 70.24 & 4.61
& 87.54 & 0.07 & 72.65 & 0.61 & 75.06 & 11.19 & 78.42 & 3.96 \\
\hline
\hline
\end{tabular}
\end{adjustbox}
\vspace{-0.4cm}
\end{table}

\begin{figure}[t]
 \setlength{\abovecaptionskip}{2pt}
  \centering
  \begin{subfigure}[t]{0.23\linewidth}
    \centering
    \includegraphics[width=\linewidth,height=0.18\textheight,keepaspectratio]{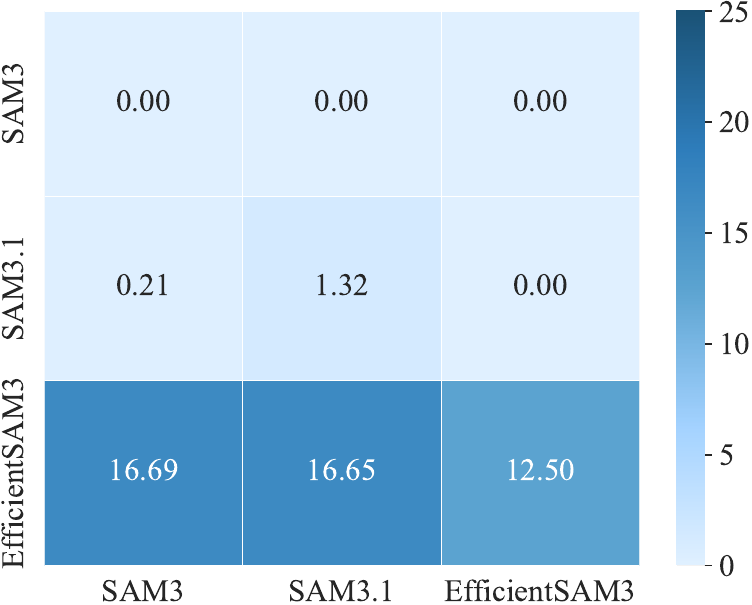}
    \caption{Text transfer}
    \label{fig:row-text-transfer}
  \end{subfigure}\hfill
  \begin{subfigure}[t]{0.245\linewidth}
    \centering
    \includegraphics[width=\linewidth,height=0.18\textheight,keepaspectratio]{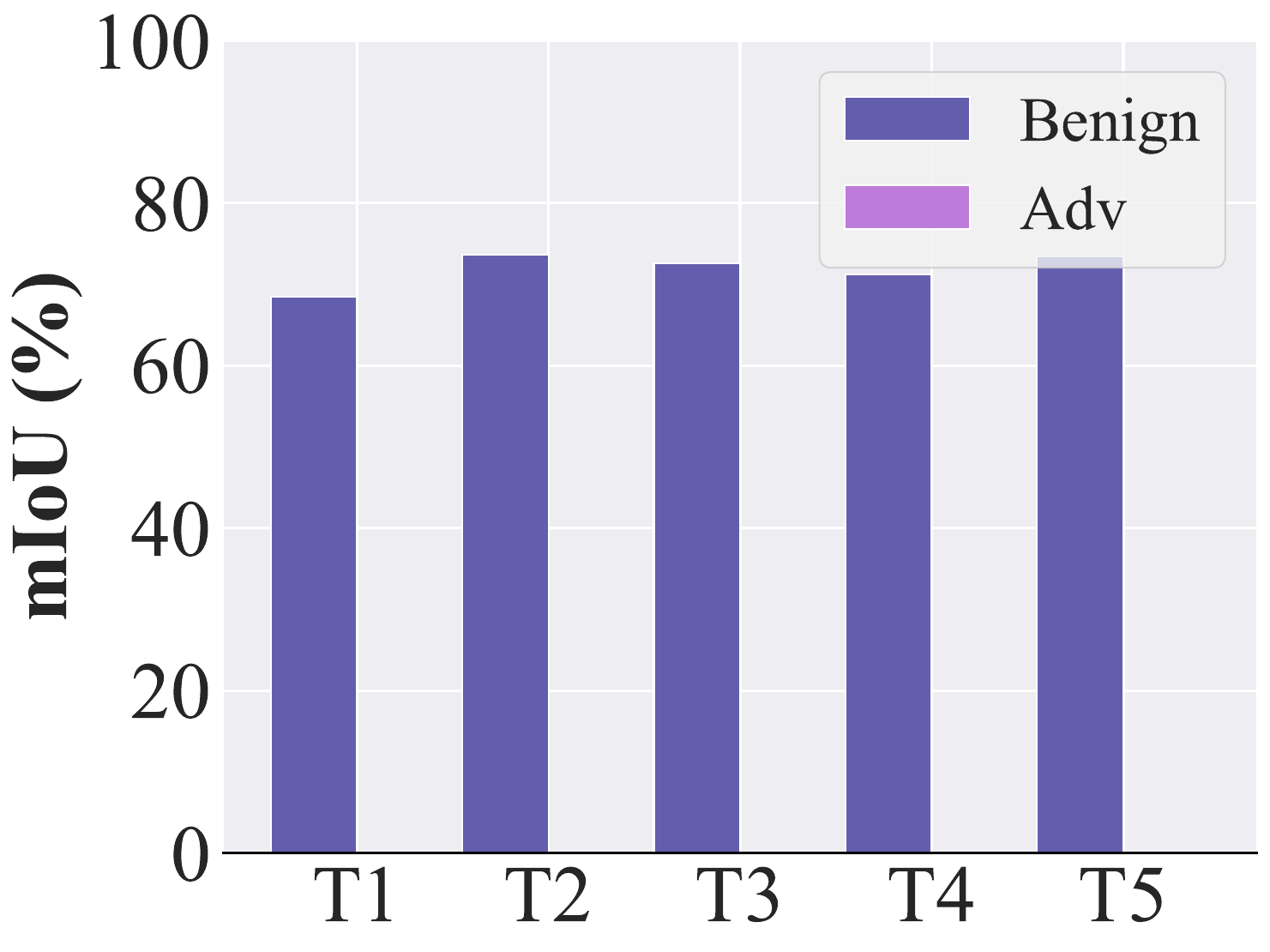}
    \caption{Text prompt}
    \label{fig:row-text-bars}
  \end{subfigure}\hfill
  \begin{subfigure}[t]{0.245\linewidth}
    \centering
\includegraphics[width=\linewidth,height=0.18\textheight,keepaspectratio]{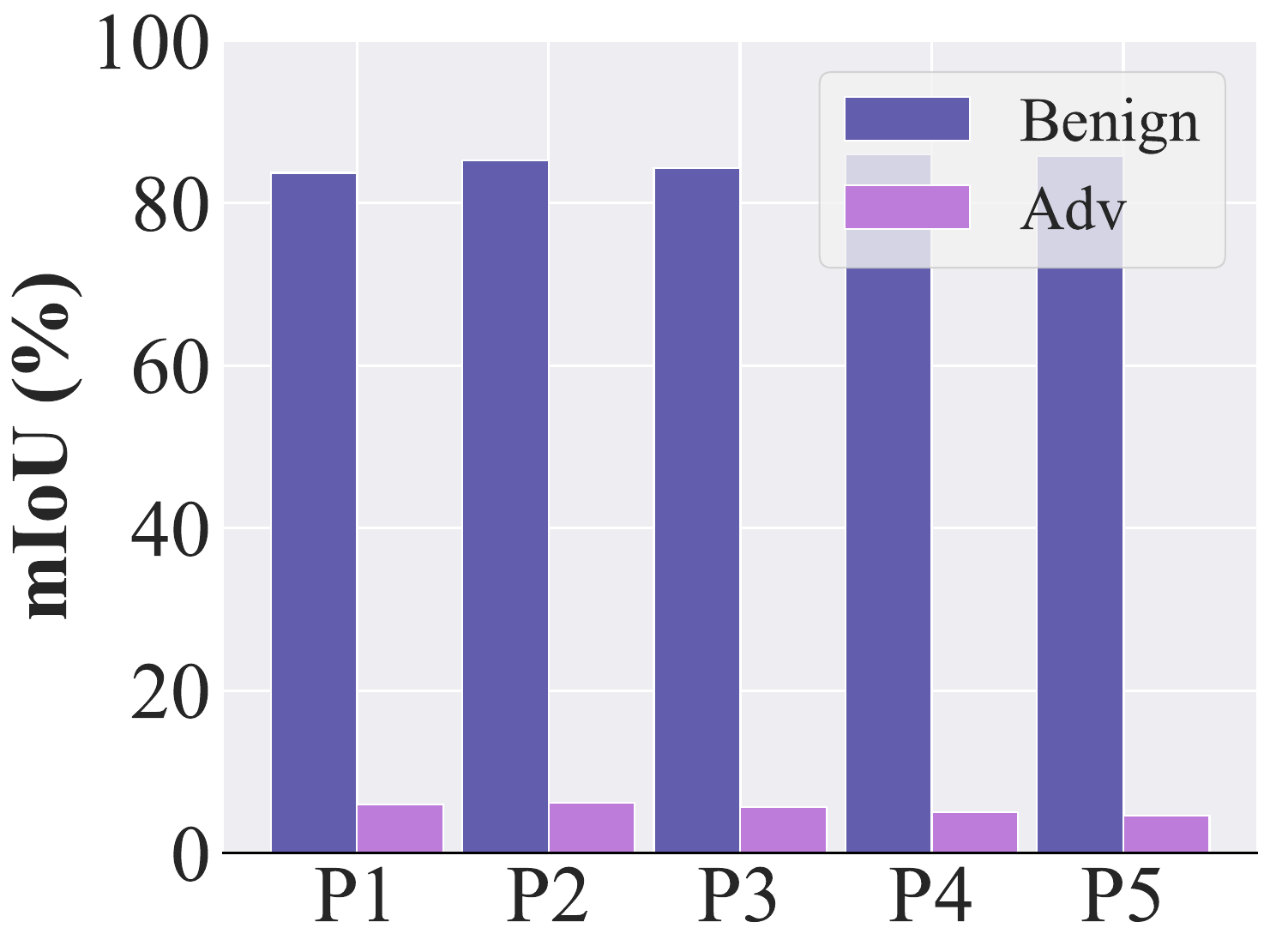}
    \caption{Point prompt}
    \label{fig:row-point-bars}
  \end{subfigure}\hfill
  \begin{subfigure}[t]{0.245\linewidth}
    \centering
\includegraphics[width=\linewidth,height=0.18\textheight,keepaspectratio]{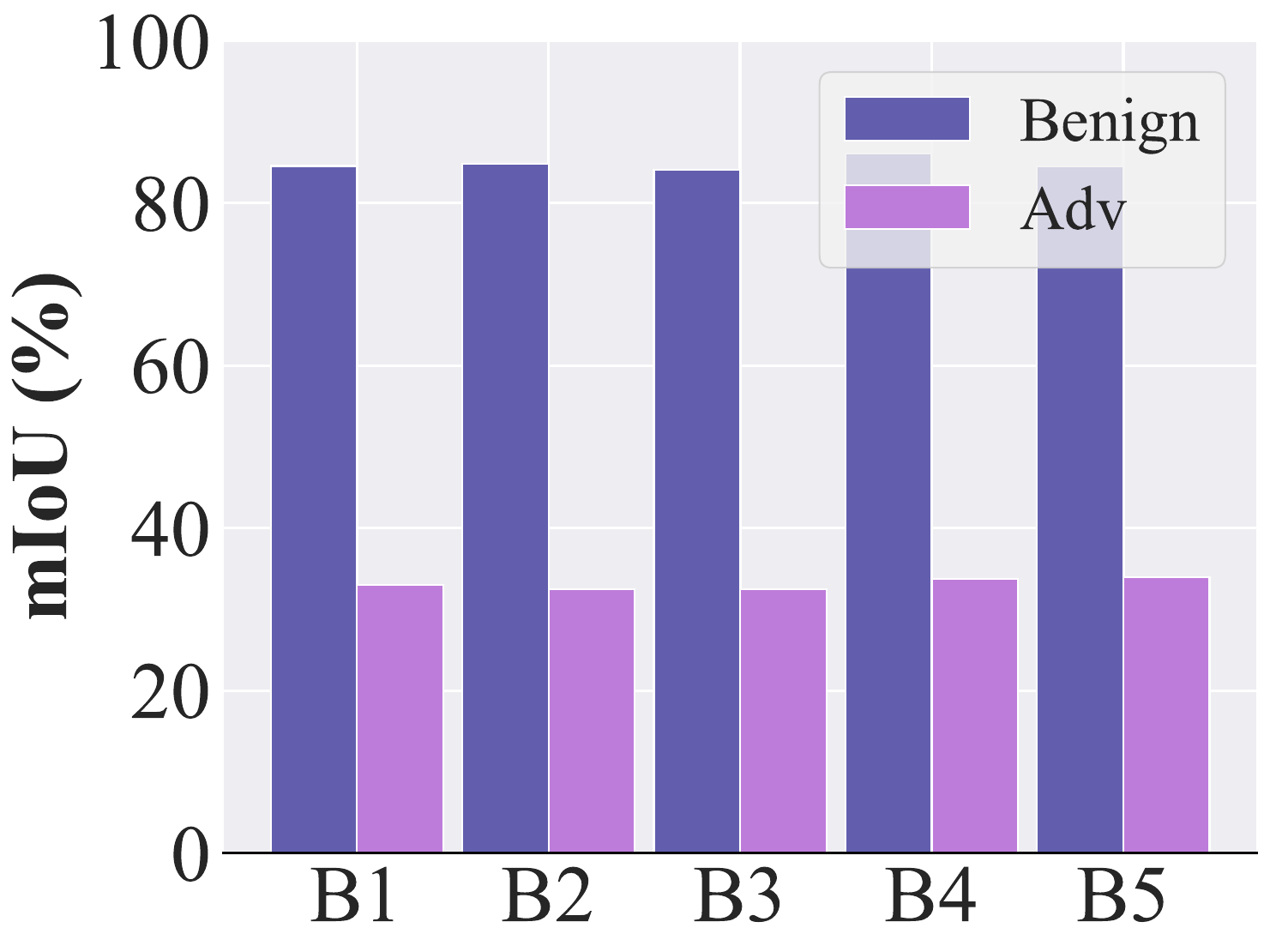}
    \caption{Box prompt}
    \label{fig:row-box-bars}
  \end{subfigure}
  \caption{Transferability study results. (a) Cross-model attack transfer results; (b)–(d) cross-prompt instance attack transfer results under different prompt types, respectively.}
  \label{fig:four-panel-separate-row}
  \vspace{-0.4cm}
\end{figure}

\noindent\textbf{Attack settings.} 
Following~\cite{zhou2024darksam,zhou2025sam2}, we set the perturbation upper bound for AdvPCS to $10/255$, with a batch size of $1$ and $15$ training epochs.
We fix the random seed to $30$ in all experiments for reproducibility.
We set $\lambda$ to $5$ and $\tau$ to $\log(0.01/0.99)$.
To fully evaluate the attack performance of AdvPCS under different prompts, we test it on SAM3 and its variants with point, box, and text prompt modes.
We evaluate AdvPCS under point, box, and text prompts on SAM3 and its variants. To better assess transferability, we train all methods on the SA-CO dataset and test them on other datasets. For video segmentation, we optimize on $100$ randomly sampled videos with $15$ frames each. For image segmentation, we use $1500$ frames for optimization.

\noindent\textbf{Evaluation metrics.} 
We use the \textit{mean Intersection over Union} (mIoU) to evaluate performance. 
It measures the average overlap between predicted masks and ground-truth masks. 
A lower mIoU indicates a stronger attack. 
We denote the mIoU computed on adversarial inputs as \textit{Adversarial mIoU} (AoU) and the mIoU computed on benign inputs as \textit{benign mIoU} (BoU).

\subsection{Attack Performance}
In this section, we evaluate the attack performance of AdvPCS on both video and image segmentation tasks, analyzing its effectiveness across datasets, models, and prompt types. 
We report quantitative results in Table~\ref{tab:main_results}. 
All UAPs are trained only on SA-CO and directly transferred to other datasets and settings. 
From the overall results, adversarial inputs consistently lead to a significant drop in mIoU across all datasets, models, and prompt types. This trend holds in both video and image settings, demonstrating the strong effectiveness of the proposed method. 
For instance, in video tasks under text prompts, the average mIoU drops sharply from $70.24\%$ to $4.61\%$ on SA-CO, and remains low on other settings, such as $17.48\%$ on YouTube and $25.90\%$ on DAVIS. 
A similar trend is observed in image settings, confirming strong cross-dataset transferability.
The attack is effective across all prompt types, including point, box, and text. Text prompts exhibit the most pronounced degradation, likely due to their higher reliance on high-level semantic reasoning. 
On the SA-CO dataset, the image mIoU drops substantially from $78.42\%$ to $3.96\%$, indicating that the model nearly fails to segment the target under adversarial perturbations. Point and box prompts also show clear performance degradation, but their AoU values remain relatively higher. 
In addition to cross-dataset transferability, as illustrated in~\cref{fig:four-panel-separate-row}, we further evaluate the transferability of AdvPCS across different SAM-series models and prompt instances. ~\cref{fig:four-panel-separate-row}(a) shows the text transfer grid from SA-CO to other models, indicating that adversarial examples generated for one model can also degrade the segmentation performance of other models, demonstrating notable cross-model transferability. ~\cref{fig:four-panel-separate-row}(b)-(d) show the attack performance on five selected groups under text, point, and box prompts, respectively. For each prompt type, all prompt instances experience consistently strong attack effects, indicating that AdvPCS maintains high cross-prompt transferability across different input prompts.

\begin{table}[t]
\centering
\scriptsize
\setlength{\tabcolsep}{2.2pt}
\renewcommand{\arraystretch}{1.4}
\caption{The mIoU (\%) of comparison study. The best results are highlighted in bold.
} 
\label{tab:comparison_video_image}
\begin{adjustbox}{center,max width=\linewidth}
\begin{tabular}{c || c|cc|cc|cc|c || c|cc|cc|cc|c}
\hline\hline
\rowcolor{gray!20}
\rule{0pt}{2.8ex} &
\multicolumn{8}{c||}{Video Segmentation} &
\multicolumn{8}{c}{Image Segmentation} \\
\rowcolor{gray!20}
&
\multicolumn{1}{c|}{SA-CO} & \multicolumn{2}{c|}{YouTube} & \multicolumn{2}{c|}{MOSE} & \multicolumn{2}{c|}{DAVIS} & Avg
& \multicolumn{1}{c|}{SA-CO} & \multicolumn{2}{c|}{YouTube} & \multicolumn{2}{c|}{MOSE} & \multicolumn{2}{c|}{DAVIS} & Avg \\
\rowcolor{gray!20}
\multirow{-3}{*}{Method}
& Text & Point & Box & Point & Box & Point & Box &
& Text & Point & Box & Point & Box & Point & Box & \\
\hline\hline
\rowcolor{gray!10}
Benign
& 68.48 & 85.29 & 88.28 & 69.12 & 75.15 & 69.73 & 88.14 & 77.74
& 87.54 & 74.17 & 88.19 & 78.90 & 86.54 & 68.21 & 87.98 & 81.65 \\
\hdashline
AttackSAM~\cite{zhang2023attack}
& 53.12 & 80.39 & 85.69 & 65.60 & 70.38 & 68.03 & 87.10 & 72.90
& 63.54 & 61.21 & 83.24 & 58.79 & 81.77 & 57.93 & 85.20 & 70.24 \\
\rowcolor{gray!10}
DarkSAM~\cite{zhou2024darksam}
& 56.16 & 83.29 & 86.97 & 66.20 & 73.57 & 62.04 & 87.44 & 73.67
& 70.22 & 64.79 & 86.68 & 65.88 & 83.81 & 61.97 & 86.78 & 74.30 \\
S-RA~\cite{shen2024practical}
& 16.74 & 32.23 & 51.18 & 34.43 & 44.16 & 34.03 & 67.40 & 40.02
& 21.63 & 11.28 & 22.59 & 21.15 & 39.74 & 17.43 & 27.22 & 23.01 \\
\rowcolor{gray!10}
UAD~\cite{lu2024unsegment}
& 6.55 & 25.79 & 50.26 & 24.14 & 33.14 & 33.12 & 53.15 & 32.31
& 7.47 & 12.43 & 48.87 & 14.08 & 46.68 & 12.67 & 47.84 & 27.15 \\
UAP-SAM2~\cite{zhou2025sam2}
& 1.18 & 22.37 & 53.96 & 13.44 & 36.62 & 24.23 & 57.01 & 29.83
& 2.86 & 4.42 & 19.21 & 3.23 & 26.41 & 4.11 & 20.84 & 11.58 \\
\hdashline
\rowcolor{gray!10}
\textbf{Ours}
& \textbf{0.00} & \textbf{6.29} & \textbf{39.62} & \textbf{8.17} & \textbf{25.20} & \textbf{7.89} & \textbf{41.13} & \textbf{18.33}
& \textbf{0.07} & \textbf{0.09} & \textbf{4.18} & \textbf{0.84} & \textbf{7.95} & \textbf{0.32} & \textbf{3.95} & \textbf{2.49} \\
\hline\hline
\end{tabular}
\end{adjustbox}
\vspace{-0.4cm}
\end{table}

\begin{figure*}[!t]   
\setlength{\abovecaptionskip}{4pt}
  \centering
     \subcaptionbox{Module}
{\vspace{2pt}\includegraphics[width=0.24\textwidth]{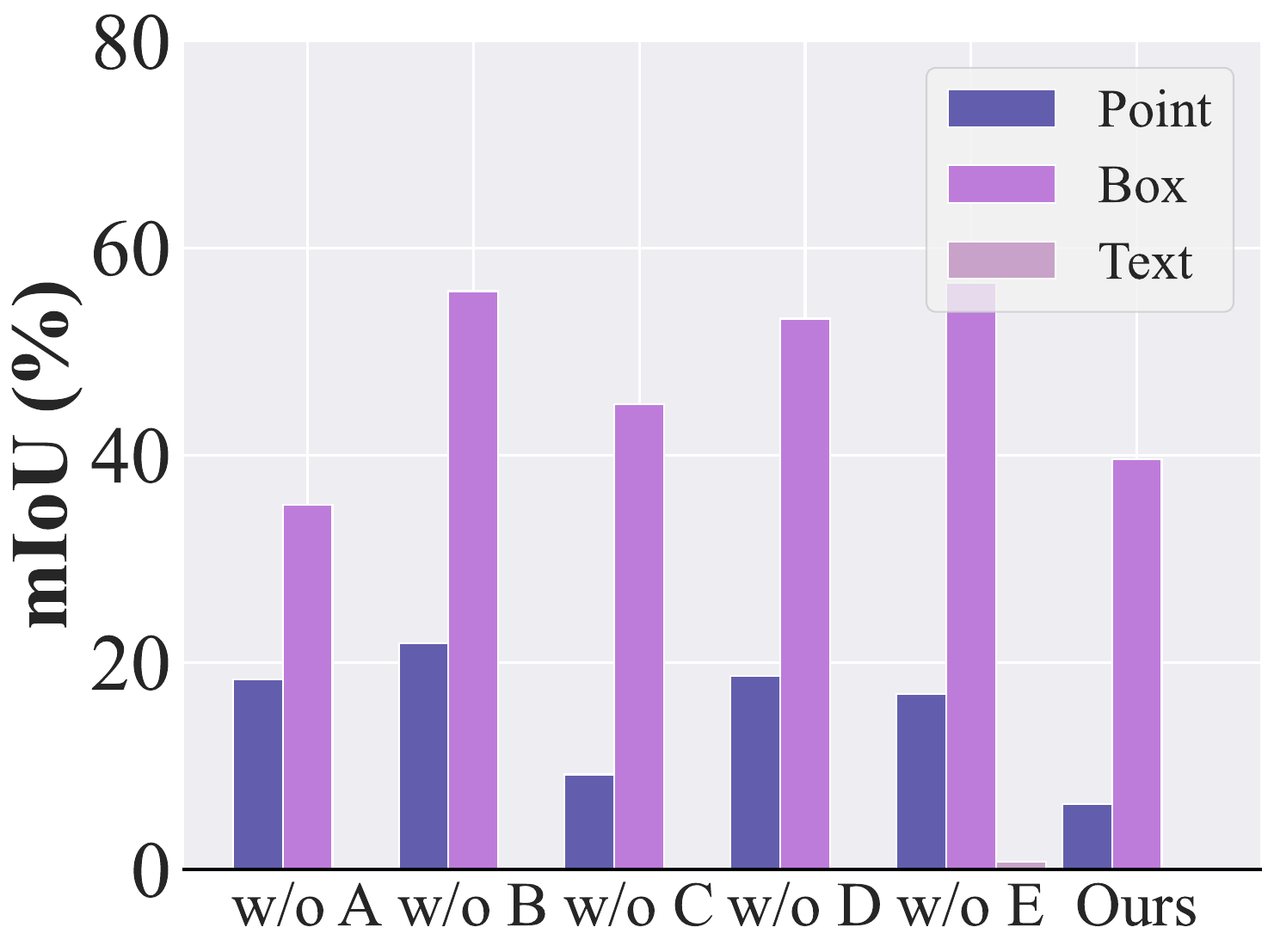}}
       \subcaptionbox{Weight}
{\vspace{2pt}\includegraphics[width=0.24\textwidth]{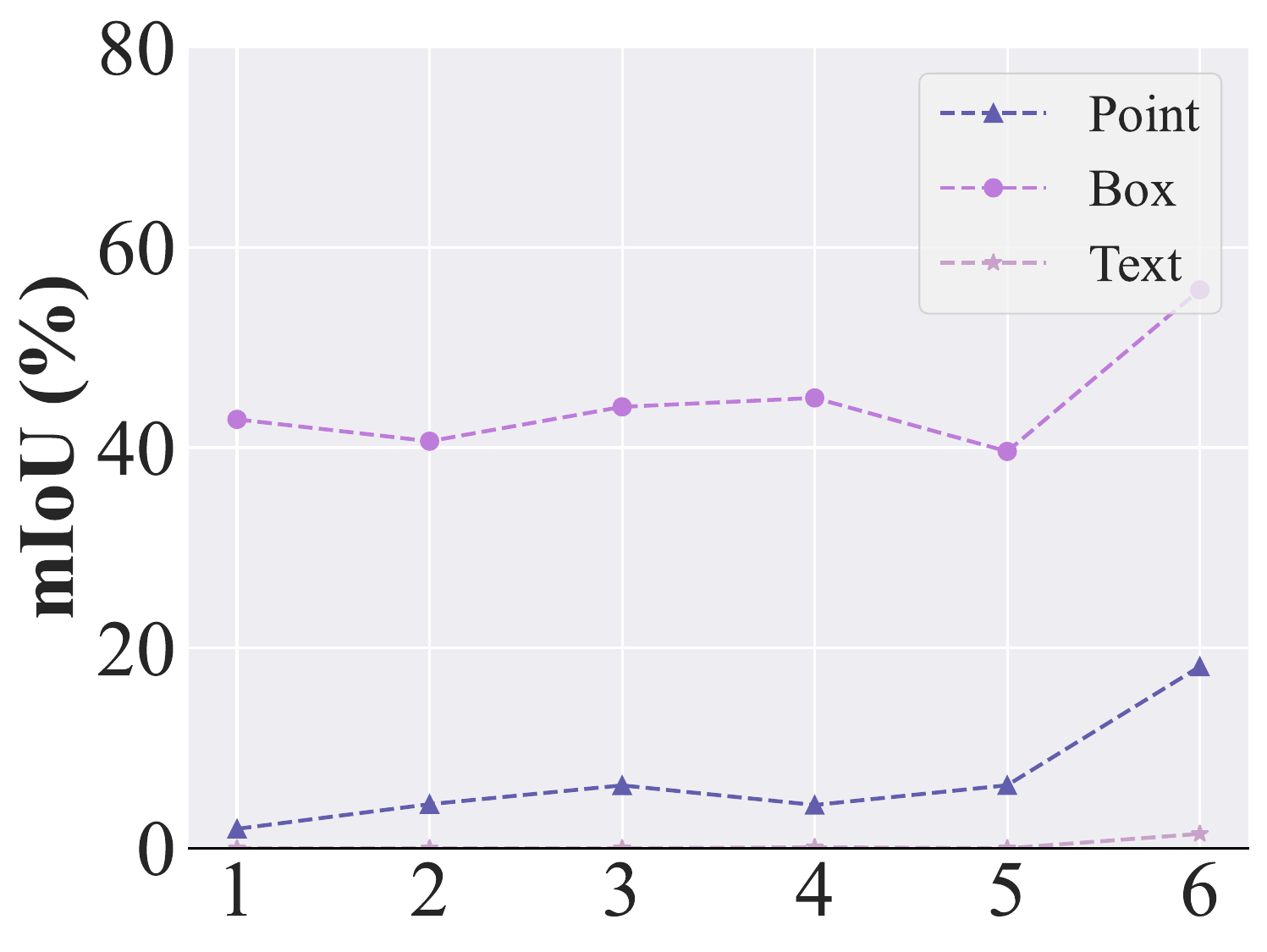}} 
        \subcaptionbox{Epoch}{\includegraphics[width=0.24\textwidth]{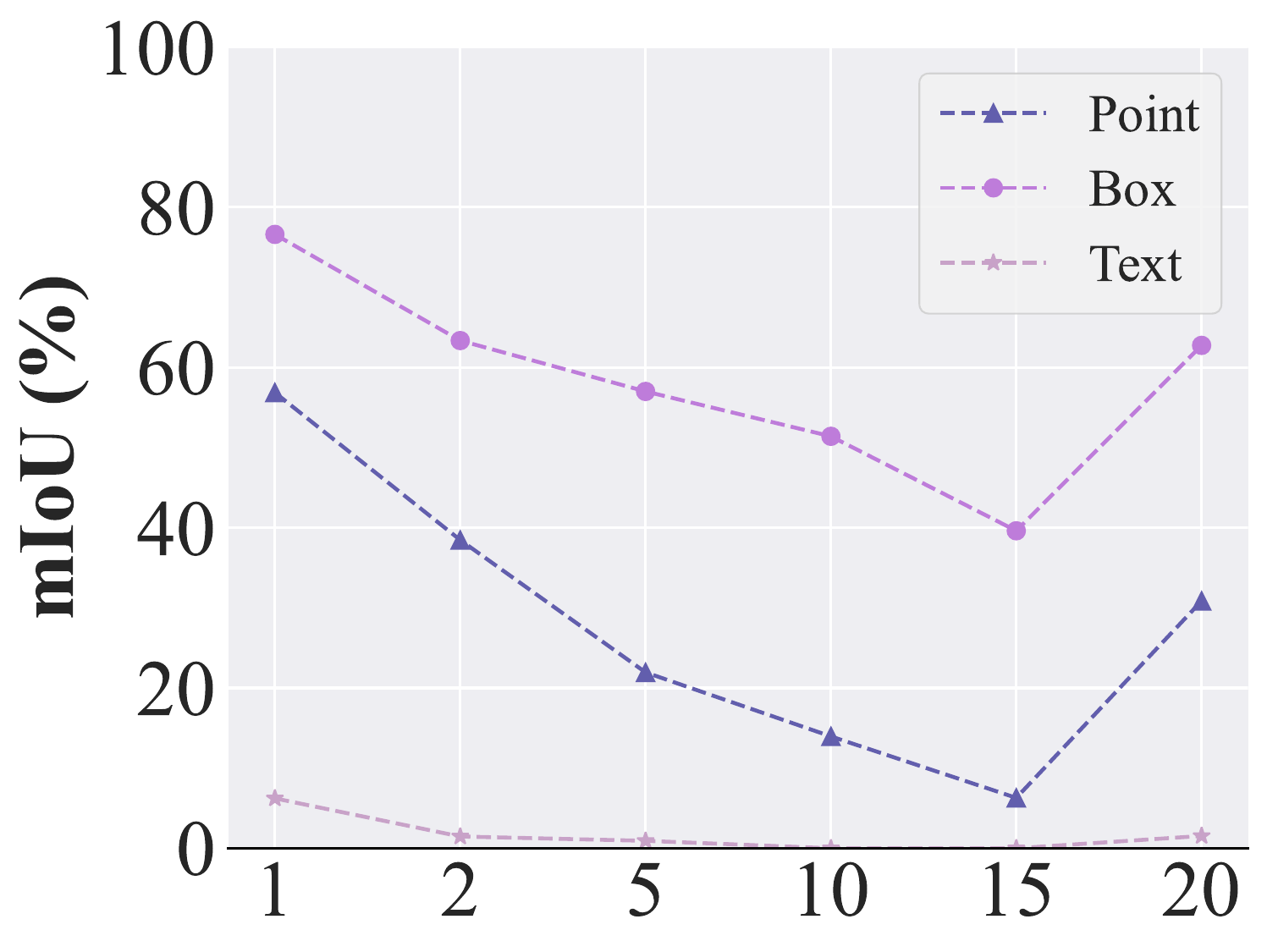}}
    \subcaptionbox{Epsilon}{\includegraphics[width=0.24\textwidth]{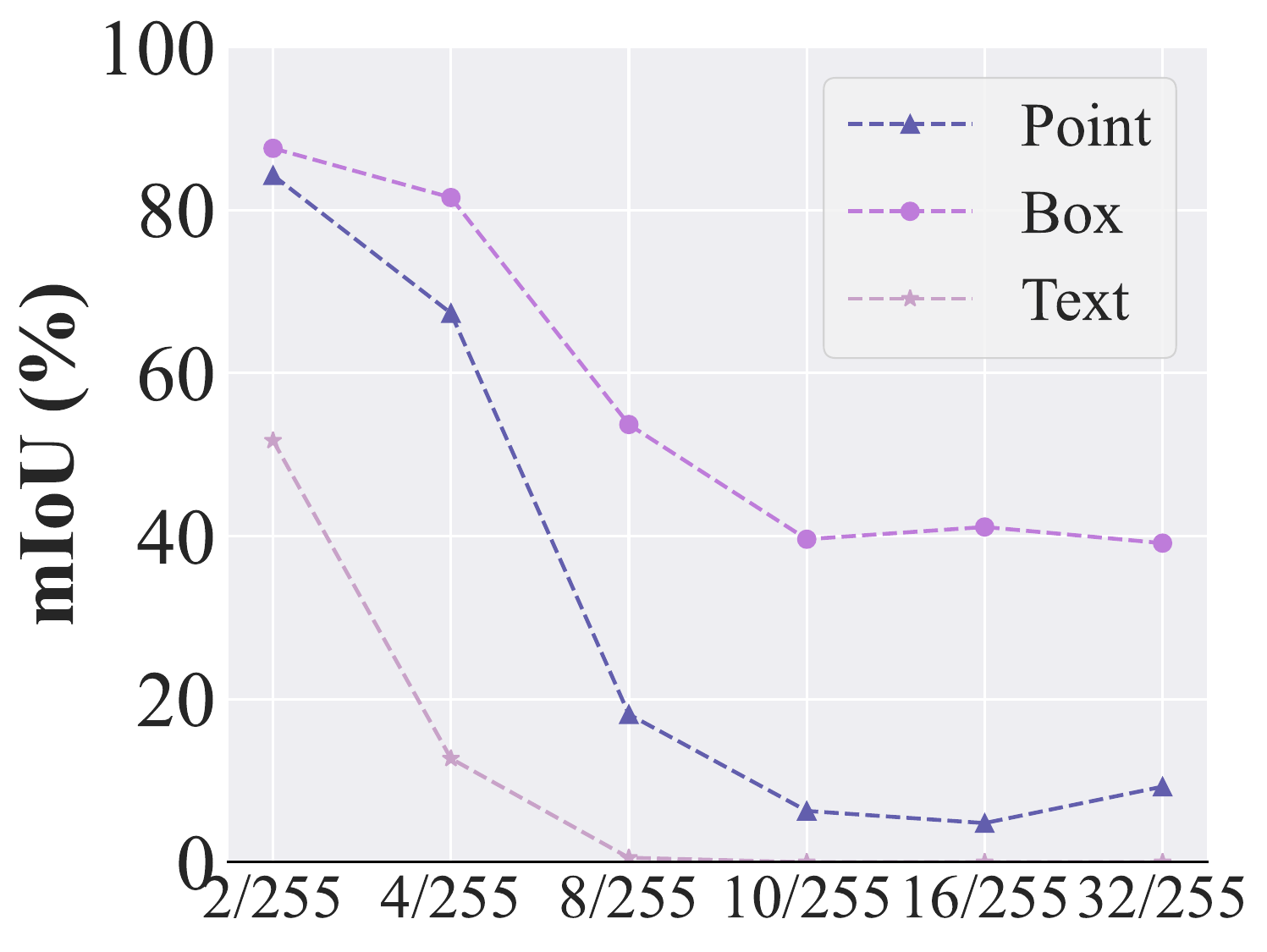}}
      \caption{Ablation results on the effect of different factors on the attack performance of AdvPCS}
       \label{fig:ablation}
       \vspace{-0.6cm}
\end{figure*}

\vspace{-0.2cm}
\subsection{Comparison Study}
In this section, we compare AdvPCS with SOTA methods, including Attack-SAM~\cite{zhang2023attack}, DarkSAM~\cite{zhou2024darksam}, S-RA~\cite{shen2024practical}, UAD~\cite{lu2024unsegment}, and UAP-SAM2~\cite{zhou2025sam2}. All baselines are adapted to a universal adversarial attack framework with the same optimization settings to ensure fair comparison. 
Table~\ref{tab:comparison_video_image} reports the quantitative results on both video and image segmentation tasks. Across all datasets, AdvPCS consistently achieves the lowest mIoU values, indicating the strongest attack performance. For instance, in video segmentation, AdvPCS reduces the average mIoU to $18.33\%$, whereas the second-best method, UAP-SAM2, achieves $29.83\%$. On the SA-CO dataset under text prompts, AdvPCS completely eliminates segmentation performance, achieving $0.00\%$ mIoU compared with $1.18\%$ for UAP-SAM2. For image segmentation, AdvPCS also outperforms all baselines, reducing the average mIoU to $2.49\%$, far below the next best result of $11.58\%$ by UAP-SAM2. 
We also provide visual comparisons of adversarial examples generated by these attacks in \cref{sec:Comparison_Study_add}.

\vspace{-0.2cm}
\subsection{Ablation Study}\label{sec:ablation}

In this section, we explore the effect of four key factors on the attack performance of AdvPCS.
The victim model is SAM3, and all experiments are conducted on the SA-CO dataset.

\noindent\textbf{The effect of modules.} 
We explore the contribution of each component within AdvPCS to its overall attack performance. Each module is sequentially removed to evaluate its individual impact. 
For simplicity, we denote the $\mathcal{J}_{local}$, $\mathcal{J}_{global}$,  $\mathcal{J}_{ta}$, $\mathcal{J}_{reg}$, and min–max bilevel adversarial prompt optimization strategy as A–E, respectively.
The results shown in~\cref{fig:ablation}(a) indicate that removing any single module significantly reduces the attack efficacy compared with the complete AdvPCS. This demonstrates that all modules are essential for achieving optimal performance. The complete AdvPCS consistently achieves the lowest mIoU across point, box, and text prompts.

\noindent\textbf{The effect of hyperparameter $\lambda$.} 
We study the effect of the weight $\lambda$ assigned to box prompts on attack performance. 
Experiments are conducted with box prompt weights ranging from $1$ to $6$. As shown in~\cref{fig:ablation}(b), attack performance increases with higher weights and reaches a peak at a weight of $5$, beyond which further increases yield minimal improvements. Based on these results, a weight of $5$ is adopted as the default configuration.

\noindent\textbf{The effect of training epochs.} 
We examine the effect of varying the number of training epochs, ranging from $1$ to $20$, on AdvPCS performance. ~\cref{fig:ablation}(c) shows that attack performance rises rapidly in the early epochs and stabilizes after $15$ epochs. Since additional epochs offer only marginal improvements, $15$ epochs are selected as the default setting.

\noindent\textbf{The effect of perturbation budget $\epsilon$.} 
We evaluate AdvPCS with perturbation budgets ranging from $2/255$ to $32/255$. As shown in~\cref{fig:ablation}(d), larger $\epsilon$ generally results in stronger attacks. Even with a small budget of $4/255$, AdvPCS significantly reduces the mIoU for all prompt types, demonstrating the method's effectiveness under limited perturbations.

\begin{figure}[t]
 \setlength{\abovecaptionskip}{2pt}
  \centering
  \begin{subfigure}[t]{0.245\linewidth}
    \centering
\includegraphics[width=\linewidth,height=0.18\textheight,keepaspectratio]{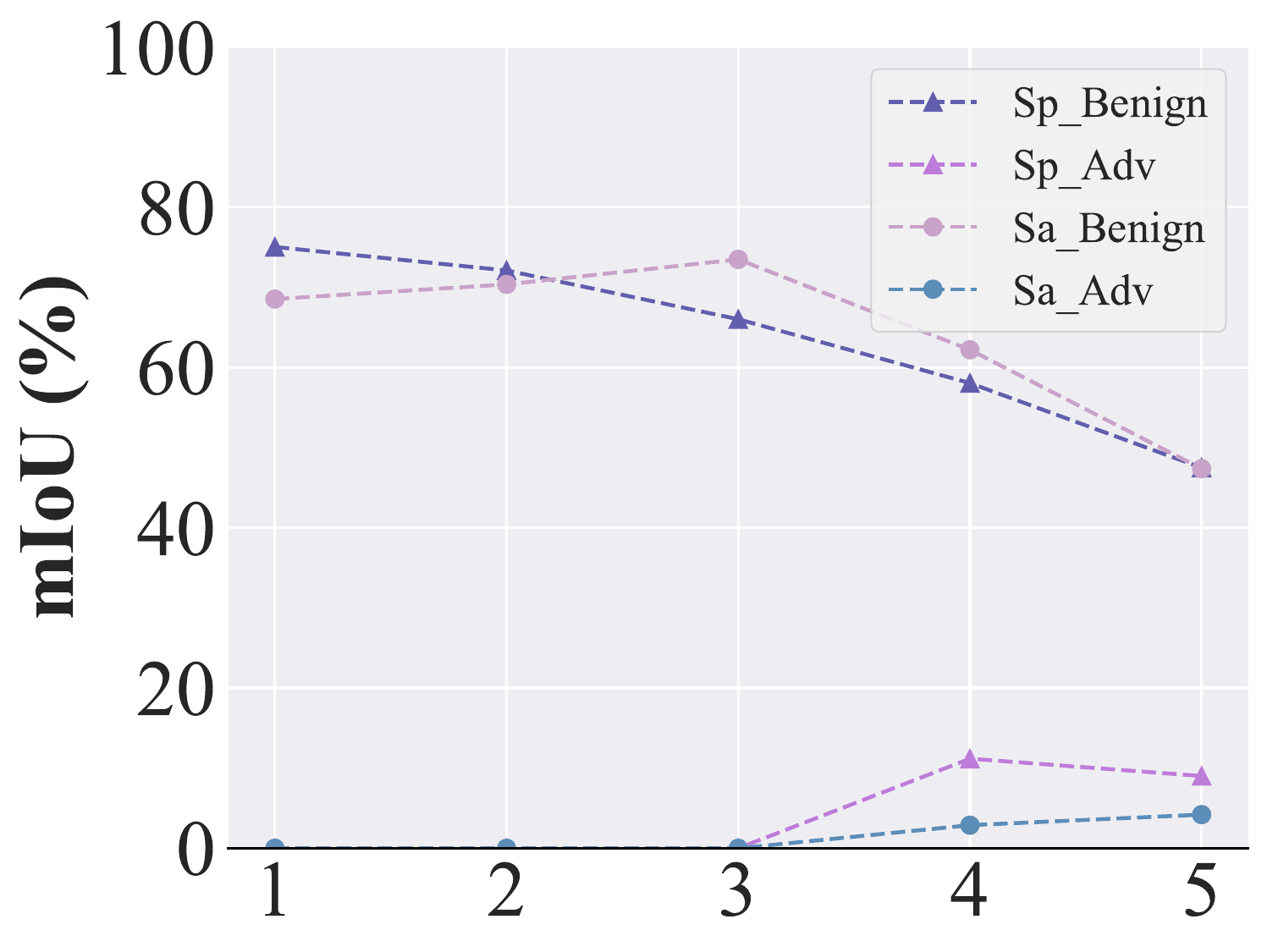}
    \caption{SAM3-Corruption}
    \label{fig:row-text-transfer}
  \end{subfigure}\hfill
\begin{subfigure}[t]{0.245\linewidth}
    \centering
\includegraphics[width=\linewidth,height=0.18\textheight,keepaspectratio]{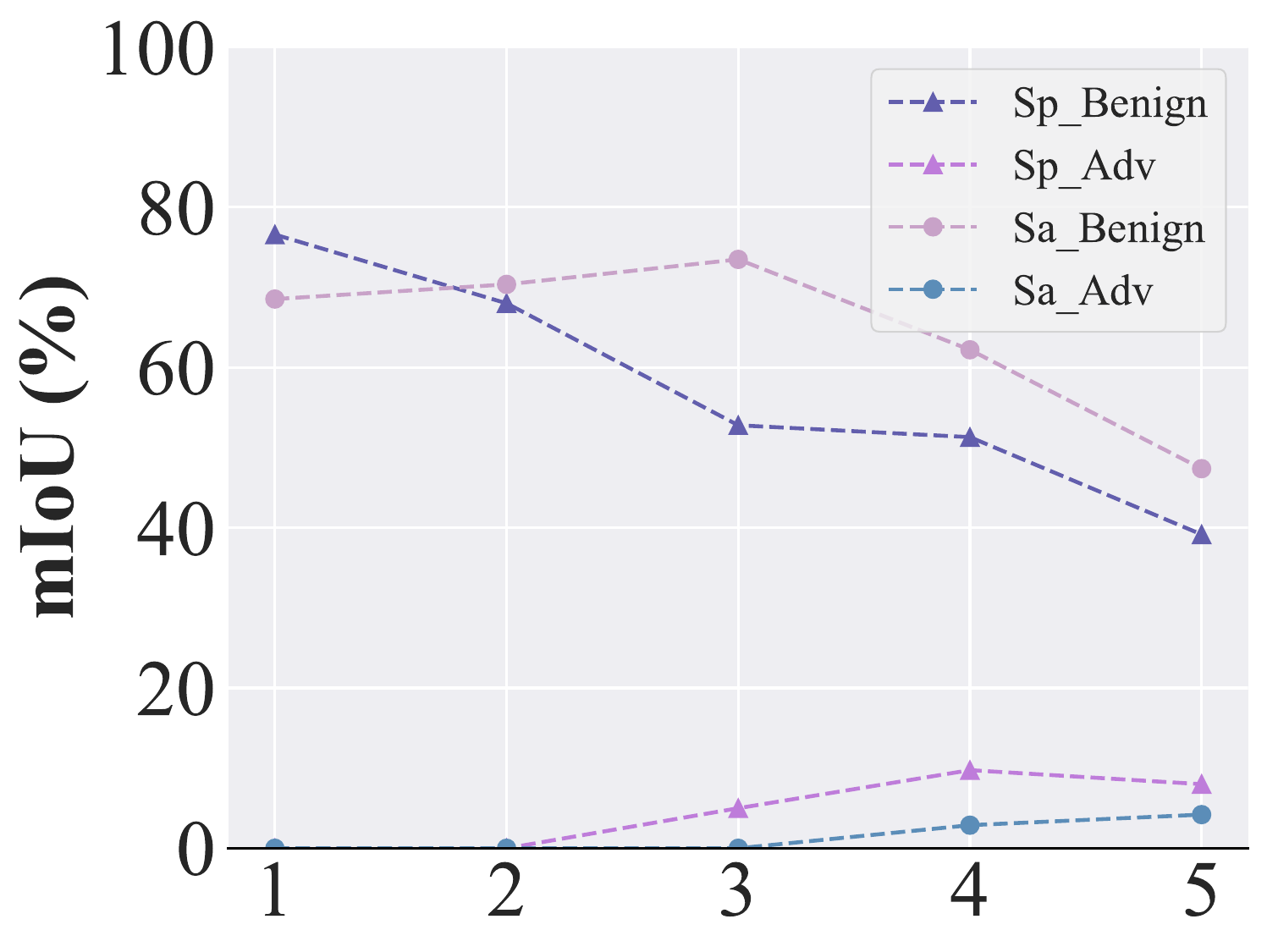}
    \caption{SAM3.1-Corruption}
    \label{fig:row-point-bars}
  \end{subfigure}\hfill
  \begin{subfigure}[t]{0.245\linewidth}
    \centering
\includegraphics[width=\linewidth,height=0.18\textheight,keepaspectratio]{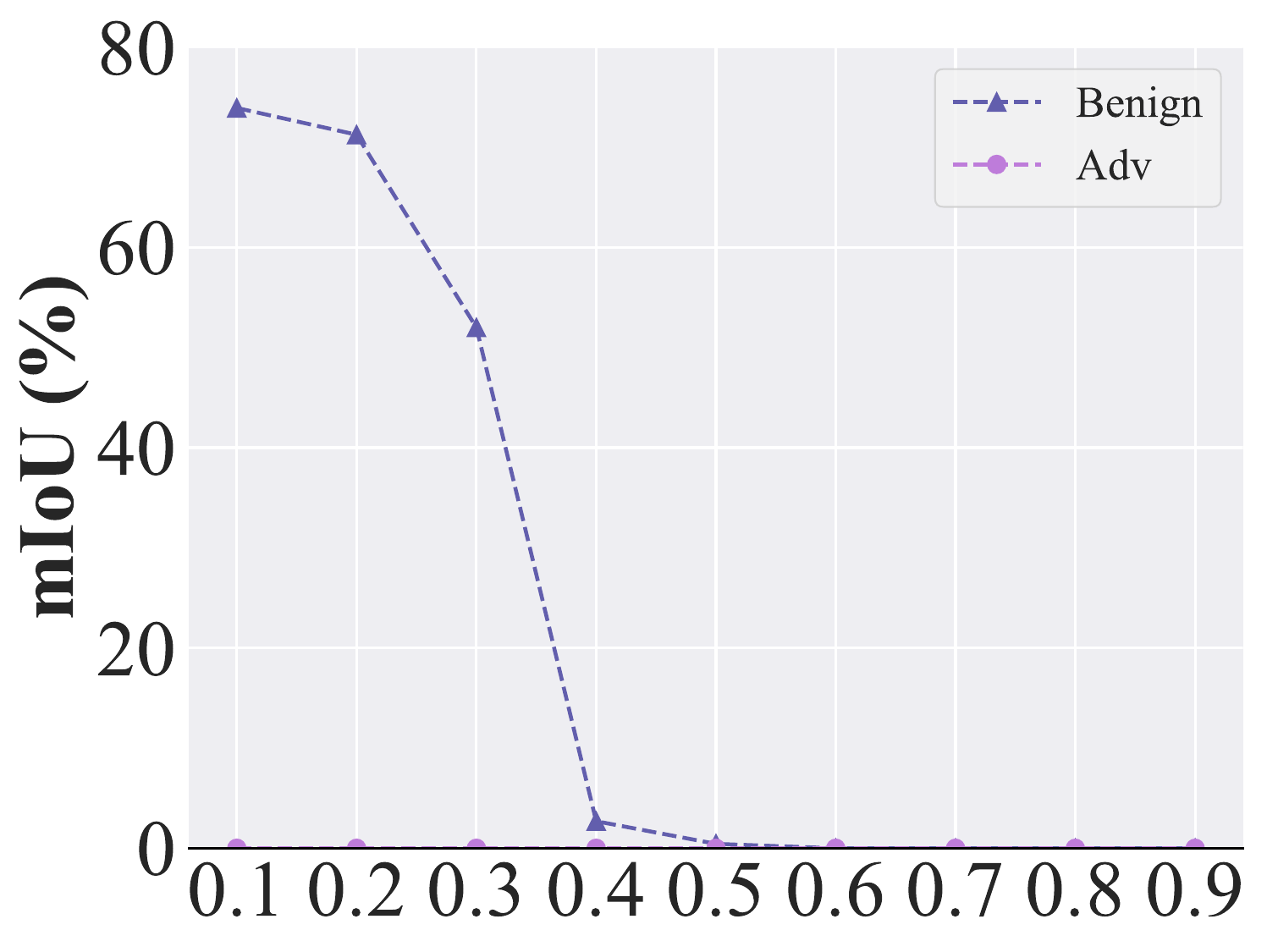}
    \caption{SAM3-Pruning}
    \label{fig:row-text-bars}
  \end{subfigure}\hfill
  \begin{subfigure}[t]{0.245\linewidth}
    \centering
\includegraphics[width=\linewidth,height=0.18\textheight,keepaspectratio]{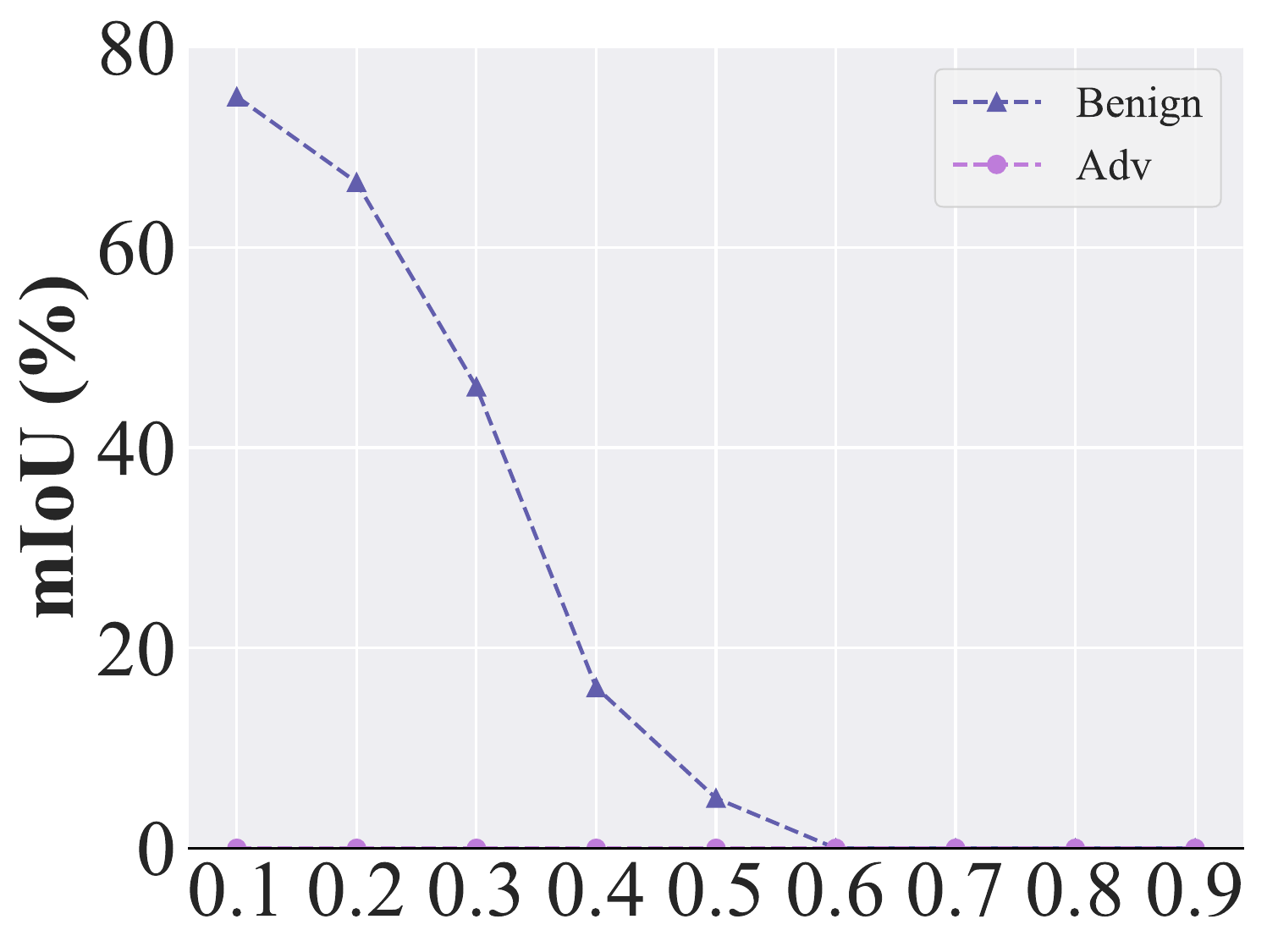}
    \caption{SAM3.1-Pruning}
    \label{fig:row-box-bars}
  \end{subfigure}
  \caption{Results of defense evaluation. (a)-(b): Data pre-processing, (c)-(d): Model pruning}
  \label{fig:defense}
  \vspace{-0.6cm}
\end{figure}

\subsection{Defense Study}
\label{sec:Defense}
In this section, we evaluate the robustness of AdvPCS against common defense strategies applied to SAM3-series models on SA-CO, including data pre-processing~\cite{prakash2018deflecting} and model pruning~\cite{zhu2017prune}.

\noindent\textbf{Data pre-processing.} 
We apply the Spatter (Sp) and Saturate (Sa) corruptions at severity levels from $0$ to $5$ on both SAM3 and SAM3.1. 
As shown in ~\cref{fig:defense}(a) and \cref{fig:defense}(b), increasing the corruption severity causes the mIoU of benign samples to drop from approximately $70\%$ to $40\%$, while the mIoU of adversarial examples remains consistently below $15\%$. This indicates that AdvPCS can still effectively degrade model performance even under strong pre-processing defenses.

\noindent\textbf{Model pruning.}
We evaluate AdvPCS under varying pruning ratios on SAM3 and SAM3.1 using the DAVIS dataset. ~\cref{fig:defense}(c) and \cref{fig:defense}(d) show that as the pruning ratio increases from $0$ to $0.6$, the mIoU of benign samples drops sharply from approximately $70\%$ to below $5\%$, whereas the mIoU of adversarial examples remains close to $0\%$. These results suggest that model pruning provides very limited effectiveness against AdvPCS.

\vspace{-0.4cm}
\section{Conclusions, Limitations, and Broader Impact}
\label{sec:Conclusion}
\vspace{-0.2cm}
In this paper, we present AdvPCS, a novel cross-prompt universal adversarial attack against PCS models.
By explicitly targeting both cross-prompt transferability and perception-level distortion, AdvPCS generates a highly transferable universal adversarial perturbation that generalizes across diverse prompts and frames. 
AdvPCS integrates a min--max bilevel adversarial prompt optimization strategy, a global–local perception deception attack that suppresses global and local presence probabilities, and a temporal transition deviation attack that disrupts semantic consistency across frames. 
Extensive experiments on four datasets and three models demonstrate that AdvPCS consistently outperforms SOTA attacks, revealing critical vulnerabilities in emerging PCS models.

Despite its effectiveness, our method has two limitations. First, the min--max bilevel optimization incurs additional computational overhead due to prompt exploration and selection. Second, AdvPCS specifically targets concept-level presence prediction in PCS models, and may not generalize to conventional segmentation models or earlier frameworks such as SAM and SAM2.
This work exposes security risks in emerging PCS models, particularly adversarial examples with cross-prompt and cross-frame transferability, and highlights the potential misuse of adversarial attacks.

 \section*{Acknowledgements}
Shengshan Hu's work is supported by the National Natural Science Foundation of China under Grant No.62372196.
Yufei Song is the is the corresponding author.

\bibliographystyle{iccv}
\bibliography{main}

\begin{thebibliography}{10}\itemsep=-1pt

\bibitem{camarena2025ad}
Mario Camarena, Het Patel, Fatemeh Nazari, Evangelos Papalexakis, Mohamadhossein Noruzoliaee, and Jia Chen.
\newblock Ad-sam: Fine-tuning the segment anything vision foundation model for autonomous driving perception.
\newblock {\em arXiv preprint arXiv:2510.27047}, 2025.

\bibitem{carion2025sam}
Nicolas Carion, Laura Gustafson, Yuan-Ting Hu, Shoubhik Debnath, Ronghang Hu, Didac Suris, Chaitanya Ryali, Kalyan~Vasudev Alwala, Haitham Khedr, Andrew Huang, et~al.
\newblock Sam 3: Segment anything with concepts.
\newblock In {\em Proceedings of The fourteen International Conference on Learning Representations (ICLR'26)}, 2026.

\bibitem{chen2023sam}
Tianrun Chen, Lanyun Zhu, Chaotao Deng, Runlong Cao, Yan Wang, Shangzhan Zhang, Zejian Li, Lingyun Sun, Ying Zang, and Papa Mao.
\newblock Sam-adapter: Adapting segment anything in underperformed scenes.
\newblock In {\em Proceedings of the IEEE/CVF International Conference on Computer Vision (ICCV'23)}, pages 3367--3375, 2023.

\bibitem{cheng2025ur}
Jian Cheng.
\newblock Ur-sam: Urban road segmentation in autonomous driving scenes based on sam.
\newblock In {\em Proceedings of the 3rd International Conference on Algorithm, Image Processing and Machine Vision (AIPMV'25)}, pages 126--130. IEEE, 2025.

\bibitem{cheng2023sam}
Junlong Cheng, Jin Ye, Zhongying Deng, Jianpin Chen, Tianbin Li, Haoyu Wang, Yanzhou Su, Ziyan Huang, Jilong Chen, Lei Jiang, et~al.
\newblock Sam-med2d.
\newblock {\em arXiv preprint arXiv:2308.16184}, 2023.

\bibitem{croce2024segment}
Francesco Croce and Matthias Hein.
\newblock Segment (almost) nothing: Prompt-agnostic adversarial attacks on segmentation models.
\newblock In {\em Proceedings of the IEEE Conference on Secure and Trustworthy Machine Learning (SaTML'24)}, pages 425--442. IEEE, 2024.

\bibitem{ding2023mose}
Henghui Ding, Chang Liu, Shuting He, Xudong Jiang, Philip H.~S. Torr, and Song Bai.
\newblock Mose: A new dataset for video object segmentation in complex scenes.
\newblock In {\em Proceedings of the IEEE/CVF International Conference on Computer Vision (ICCV'23)}, pages 20224--20234. IEEE, 2023.

\bibitem{han2023sam}
Dongshen Han, Chaoning Zhang, Sheng Zheng, Chang Lu, Yang Yang, and Heng~Tao Shen.
\newblock Sam meets uap: Attacking segment anything model with universal adversarial perturbation.
\newblock {\em arXiv preprint arXiv:2310.12431}, 2023.

\bibitem{huang2024segment}
Shize Huang, Qianhui Fan, Zhaoxin Zhang, Xiaowen Liu, Guanqun Song, and Jinzhe Qin.
\newblock Segment shards: Cross-prompt adversarial attacks against the segment anything model.
\newblock {\em Applied Sciences}, 14(8):3312, 2024.

\bibitem{jiang2024cross}
Yan Jiang, Guisheng Yin, Ye Yuan, Jingjing Chen, and Zhipeng Wei.
\newblock Cross-point adversarial attack based on feature neighborhood disruption against segment anything model.
\newblock In {\em Proceedings of the IEEE International Conference on Multimedia and Expo (ICME'25)}, pages 1--6. IEEE, 2024.

\bibitem{ke2023segment}
Lei Ke, Mingqiao Ye, Martin Danelljan, Yu-Wing Tai, Chi-Keung Tang, Fisher Yu, et~al.
\newblock Segment anything in high quality.
\newblock In {\em Proceedings of the 37th Annual Conference on Neural Information Processing Systems (NeurIPS'23)}, volume~36, pages 29914--29934, 2023.

\bibitem{kirillov2023segment}
Alexander Kirillov, Eric Mintun, Nikhila Ravi, Hanzi Mao, Chlo{\'{e}} Rolland, Laura Gustafson, Tete Xiao, Spencer Whitehead, Alexander~C. Berg, Wan{-}Yen Lo, Piotr Doll{\'{a}}r, and Ross~B. Girshick.
\newblock Segment anything.
\newblock In {\em Proceedings of the IEEE/CVF International Conference on Computer Vision (ICCV'23)}, pages 3992--4003. IEEE, 2023.

\bibitem{kweon2024sam}
Hyeokjun Kweon and Kuk-Jin Yoon.
\newblock From sam to cams: Exploring segment anything model for weakly supervised semantic segmentation.
\newblock In {\em Proceedings of the IEEE/CVF Conference on Computer Vision and Pattern Recognition (CVPR'24)}, pages 19499--19509, 2024.

\bibitem{liu2024cross}
Yuchen Liu and Pengxu Wei.
\newblock Cross-prompt adversarial attack on segment anything model.
\newblock In {\em Proceedings of the 12th International Conference on Communications and Broadband Networking (ICCBN'24)}, pages 34--39, 2024.

\bibitem{long2025robust}
Jiahuan Long, Zhengqin Xu, Tingsong Jiang, Wen Yao, Shuai Jia, Chao Ma, and Xiaoqian Chen.
\newblock Robust sam: on the adversarial robustness of vision foundation models.
\newblock In {\em Proceedings of the AAAI Conference on Artificial Intelligence (AAAI'25)}, volume~39, pages 5775--5783, 2025.

\bibitem{lu2024unsegment}
Jiahao Lu, Xingyi Yang, and Xinchao Wang.
\newblock Unsegment anything by simulating deformation.
\newblock In {\em Proceedings of the IEEE/CVF Conference on Computer Vision and Pattern Recognition (CVPR'24)}, pages 24294--24304, 2024.

\bibitem{madry2017towards}
Aleksander Madry, Aleksandar Makelov, Ludwig Schmidt, Dimitris Tsipras, and Adrian Vladu.
\newblock Towards deep learning models resistant to adversarial attacks.
\newblock In {\em Proceedings of the International Conference on Learning Representations (ICLR'18)}, 2018.

\bibitem{pont20172017}
Jordi Pont-Tuset, Federico Perazzi, Sergi Caelles, Pablo Arbel{\'a}ez, Alex Sorkine-Hornung, and Luc Van~Gool.
\newblock The 2017 davis challenge on video object segmentation.
\newblock {\em arXiv preprint arXiv:1704.00675}, 2017.

\bibitem{prakash2018deflecting}
Aaditya Prakash, Nick Moran, Solomon Garber, Antonella DiLillo, and James Storer.
\newblock Deflecting adversarial attacks with pixel deflection.
\newblock In {\em Proceedings of the IEEE conference on computer vision and pattern recognition (CVPR'18)}, pages 8571--8580, 2018.

\bibitem{qiao2023robustness}
Yu Qiao, Chaoning Zhang, Taegoo Kang, Donghun Kim, Chenshuang Zhang, and Choong~Seon Hong.
\newblock Robustness of sam: Segment anything under corruptions and beyond.
\newblock {\em arXiv preprint arXiv:2306.07713}, 2023.

\bibitem{ravi2024sam2}
Nikhila Ravi, Valentin Gabeur, Yuan{-}Ting Hu, Ronghang Hu, Chaitanya Ryali, Tengyu Ma, Haitham Khedr, Roman R{\"{a}}dle, Chlo{\'{e}} Rolland, Laura Gustafson, Eric Mintun, Junting Pan, Alwala Kalyan~Vasudev, Nicolas Carion, Chao{-}Yuan Wu, Ross~B. Girshick, Piotr Doll{\'{a}}r, and Christoph Feichtenhofer.
\newblock Sam 2: Segment anything in images and videos.
\newblock In {\em Proceedings of The Thirteen International Conference on Learning Representations (ICLR'25)}, 2025.

\bibitem{sengupta2025sam}
Sourya Sengupta, Satrajit Chakrabarty, and Ravi Soni.
\newblock Is sam 2 better than sam in medical image segmentation?
\newblock In {\em Proceedings of the Medical Imaging 2025: Image Processing (SPIE'25)}, volume 13406, pages 666--672. SPIE, 2025.

\bibitem{shen2024practical}
Yifan Shen, Zhengyuan Li, and Gang Wang.
\newblock Practical region-level attack against segment anything models.
\newblock In {\em Proceedings of the IEEE/CVF Conference on Computer Vision and Pattern Recognition (CVPR'24)}, pages 194--203, 2024.

\bibitem{song2025seg}
Yufei Song, Ziqi Zhou, Qi Lu, Hangtao Zhang, Yifan Hu, Lulu Xue, Shengshan Hu, Minghui Li, and Leo~Yu Zhang.
\newblock Segtrans: Transferable adversarial examples for segmentation models.
\newblock {\em IEEE Transactions on Multimedia}, 2025.

\bibitem{vaswani2017attention}
Ashish Vaswani, Noam Shazeer, Niki Parmar, Jakob Uszkoreit, Llion Jones, Aidan~N Gomez, {\L}ukasz Kaiser, and Illia Polosukhin.
\newblock Attention is all you need.
\newblock In {\em Proceedings of the 31st Annual Conference on Neural Information Processing Systems (NeurIPS'17)}, volume~30, 2017.

\bibitem{wang2026x}
Hao Wang, Limeng Qiao, Zequn Jie, Zhijian Huang, Chengjian Feng, Qingfang Zheng, Lin Ma, Xiangyuan Lan, and Xiaodan Liang.
\newblock X-sam: From segment anything to any segmentation.
\newblock In {\em Proceedings of the AAAI Conference on Artificial Intelligence (AAAI'26)}, volume~40, pages 26187--26196, 2026.

\bibitem{xia2024transferable}
Song Xia, Wenhan Yang, Yi Yu, Xun Lin, Henghui Ding, Lingyu Duan, and Xudong Jiang.
\newblock Transferable adversarial attacks on sam and its downstream models.
\newblock In {\em Proceedings of the 38th Annual Conference on Neural Information Processing Systems (NeurIPS'24)}, pages 87545--87568, 2024.

\bibitem{xu2018youtube}
Ning Xu, Linjie Yang, Yuchen Fan, Dingcheng Yue, Yuchen Liang, Jianchao Yang, and Thomas Huang.
\newblock Youtube-vos: A large-scale video object segmentation benchmark.
\newblock {\em arXiv preprint arXiv:1809.03327}, 2018.

\bibitem{zeng2025efficientsam3}
Chengxi Zeng, Yuxuan Jiang, and Aaron Zhang.
\newblock Efficientsam3: Progressive hierarchical distillation for video concept segmentation from sam1, 2, and 3.
\newblock {\em arXiv preprint arXiv:2511.15833}, 2025.

\bibitem{zhang2023attack}
Chenshuang Zhang, Chaoning Zhang, Taegoo Kang, Donghun Kim, Sung-Ho Bae, and In~So Kweon.
\newblock Attack-sam: Towards evaluating adversarial robustness of segment anything model.
\newblock {\em arXiv preprint arXiv:2305.00866}, 1(3):5, 2023.

\bibitem{zhang2026efficient}
Jing Zhang, Zhikai Li, Xuewen Liu, and Qingyi Gu.
\newblock Efficient-sam2: Accelerating sam2 with object-aware visual encoding and memory retrieval.
\newblock In {\em Proceedings of the fourteen International Conference on Learning Representations (ICLR'26)}, 2026.

\bibitem{zheng2024black}
Sheng Zheng, Chaoning Zhang, and Xinhong Hao.
\newblock Black-box targeted adversarial attack on segment anything (sam).
\newblock {\em IEEE Transactions on Multimedia}, 27:1901--1913, 2024.

\bibitem{zhou2023advclip}
Ziqi Zhou, Shengshan Hu, Minghui Li, Hangtao Zhang, Yechao Zhang, and Hai Jin.
\newblock Advclip: Downstream-agnostic adversarial examples in multimodal contrastive learning.
\newblock In {\em Proceedings of the 32nd ACM International Conference on Multimedia (MM'23)}, pages 6311--6320, 2023.

\bibitem{zhou2023downstream}
Ziqi Zhou, Shengshan Hu, Ruizhi Zhao, Qian Wang, Leo~Yu Zhang, Junhui Hou, and Hai Jin.
\newblock Downstream-agnostic adversarial examples.
\newblock In {\em Proceedings of the 2023 IEEE/CVF International Conference on Computer Vision (ICCV'23)}, pages 4345--4355, 2023.

\bibitem{zhou2025sam2}
Ziqi Zhou, Yifan Hu, Yufei Song, Zijing Li, Shengshan Hu, Leo~Yu Zhang, Dezhong Yao, Long Zheng, and Hai Jin.
\newblock Vanish into thin air: Cross-prompt universal adversarial attacks for sam2.
\newblock In {\em Proceedings of the 39th Annual Conference on Neural Information Processing Systems (NeurIPS'25)}, 2025.

\bibitem{zhou2025numbod}
Ziqi Zhou, Bowen Li, Yufei Song, Shengshan Hu, Wei Wan, Leo~Yu Zhang, Dezhong Yao, and Hai Jin.
\newblock Numbod: A spatial-frequency fusion attack against object detectors.
\newblock In {\em Proceedings of the 39th Annual AAAI Conference on Artificial Intelligence (AAAI'25)}, 2025.

\bibitem{zhou2024securely}
Ziqi Zhou, Minghui Li, Wei Liu, Shengshan Hu, Yechao Zhang, Wei Wan, Lulu Xue, Leo~Yu Zhang, Dezhong Yao, and Hai Jin.
\newblock Securely fine-tuning pre-trained encoders against adversarial examples.
\newblock In {\em Proceedings of the 2024 IEEE Symposium on Security and Privacy (SP'24)}, 2024.

\bibitem{zhou2024darksam}
Ziqi Zhou, Yufei Song, Minghui Li, Shengshan Hu, Xianlong Wang, Leo~Yu Zhang, Dezhong Yao, and Hai Jin.
\newblock Darksam: Fooling segment anything model to segment nothing.
\newblock In {\em Proceedings of the 38th Annual Conference on Neural Information Processing Systems (NeurIPS'24)}, 2024.

\bibitem{zhu2017prune}
Michael Zhu and Suyog Gupta.
\newblock To prune, or not to prune: exploring the efficacy of pruning for model compression.
\newblock {\em arXiv preprint arXiv:1710.01878}, 2017.

\end{thebibliography}

\newpage
\appendix
\setcounter{table}{0}
\setcounter{figure}{0}
\renewcommand{\thetable}{A\arabic{table}}
\renewcommand{\thefigure}{A\arabic{figure}}

\section{Contents}
\begin{itemize}

\item Sec. \ref{sec:appendix-setting}: Experimental settings including datasets, evaluation metrics, and platform.


\item Sec. \ref{sec:Stability analysis}: Stability analysis under different random seeds.

\item Sec. \ref{sec:Multi-point Evaluation Study}: Multi-point evaluation study of the proposed method under multi-prompt settings.

\item Sec. \ref{sec:Comparison_Study_add}: Supplementary visualization results of the comparison study.

\end{itemize}

\section{Experimental Setting}\label{sec:appendix-setting}
In this section, we provide details of our experimental settings.

\subsection{Datasets}\label{dataset}
We evaluate our method on four benchmark datasets: YouTube-VOS~\cite{xu2018youtube}, DAVIS~\cite{pont20172017}, MOSE~\cite{ding2023mose}, and SA-CO~\cite{carion2025sam}.

\begin{itemize}
\item {\bf YouTube:} 
YouTube~\cite{xu2018youtube} is a large-scale benchmark for video object segmentation. It contains 3,471 training videos, 474 validation videos, and 508 test videos, covering 94 object categories with pixel-level segmentation annotations. The dataset is widely used for video object segmentation, semi-supervised segmentation, and multi-object tracking.

\item {\bf DAVIS:} 
DAVIS~\cite{pont20172017} is a standard benchmark for video object segmentation. The DAVIS 2017 split contains 60 training videos, 30 validation videos, and 30 test-dev videos with high-quality pixel-level annotations. It is commonly used to evaluate precise object boundary segmentation and temporal consistency in multi-object video segmentation.

\item {\bf MOSE:} 
MOSE~\cite{ding2023mose} is a large-scale benchmark for video object segmentation in complex scenes. It contains 1,507 training videos, 311 validation videos, and 331 test videos with pixel-level segmentation annotations. The dataset focuses on challenging scenarios such as heavy occlusion, object disappearance and reappearance, and frequent object interactions.

\item {\bf SA-CO:} 
SA-CO~\cite{carion2025sam} is a large-scale dataset for open-vocabulary video segmentation. Its video component contains 52.5K videos and 467K spatiotemporal mask sequences with phrase-level segmentation annotations. The dataset supports research on text-prompted video segmentation, tracking, and open-vocabulary understanding.
\end{itemize}

\subsection{Evaluation Metrics}\label{metrics}
We use \textit{mean Intersection over Union} (mIoU) to evaluate segmentation performance.
mIoU computes the average \textit{Intersection over Union} (IoU) across all categories, where IoU measures the overlap between the predicted mask and the ground-truth mask:

\begin{equation}
    \text{IoU} = \frac{\text{Predicted Region} \cap \text{Ground Truth Region}}{\text{Predicted Region} \cup \text{Ground Truth Region}}.
\nonumber
\end{equation}
mIoU is the average of the IoUs for all categories, reflecting the model's overall performance in the segmentation task. A higher mIoU indicates better segmentation performance across categories, particularly in tasks with class imbalance or fine-grained segmentation.

\subsection{Platform}\label{sec:Platform}
We conduct experiments on a server running a 64-bit Ubuntu 20.04.1 system with an Intel(R) Xeon(R) Silver 4210R CPU @ 2.40GHz processor, 125GB memory, and two NVIDIA A100-SXM4 GPUs, each with 80GB memory. The experiments are performed using the Python language and PyTorch library version 2.1.0.


\begin{wrapfigure}{r}{0.4\textwidth}  
  \centering
  \includegraphics[width=\linewidth]{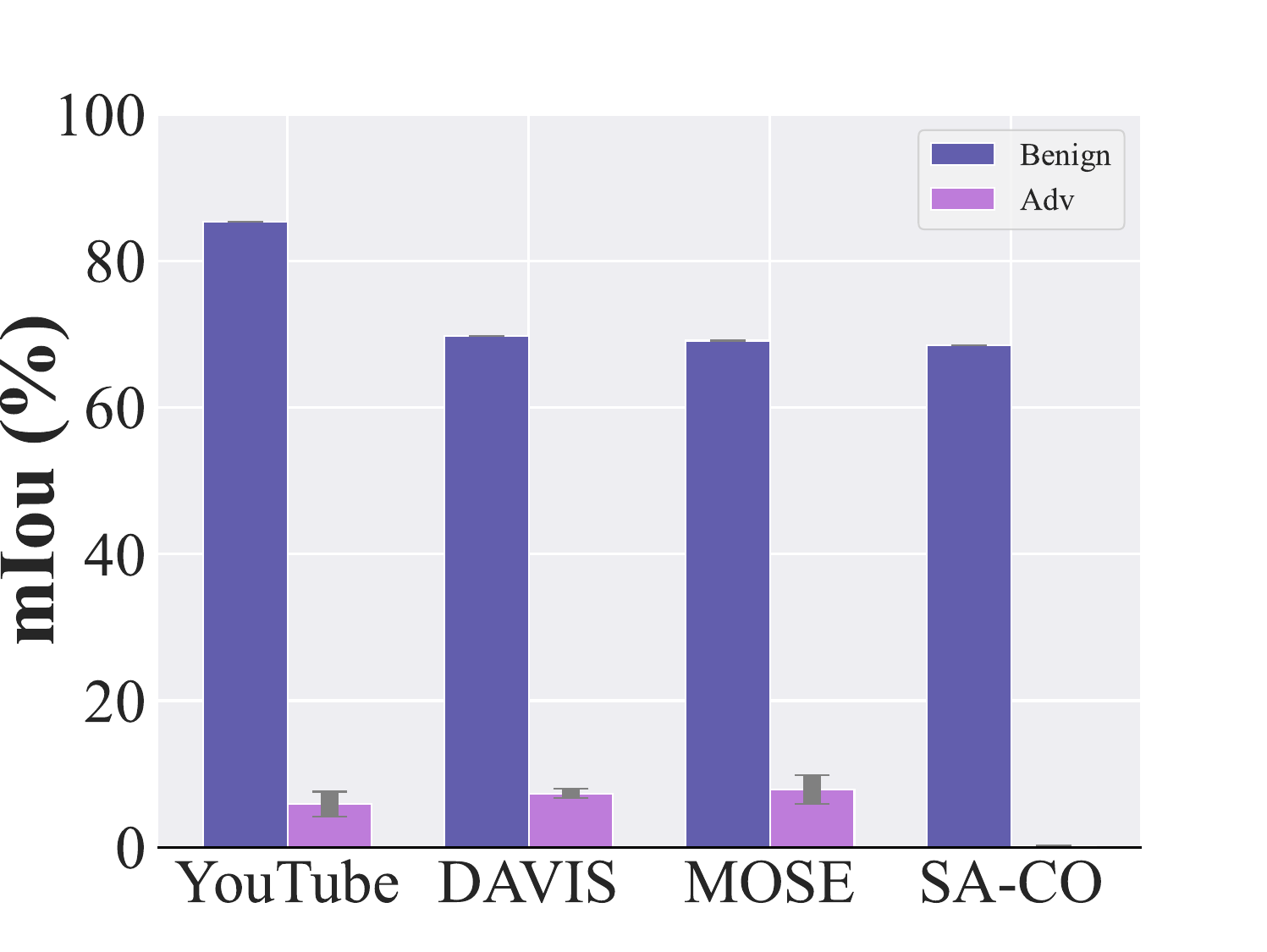}
  \caption{Stability analysis}
  \label{fig:error_bar}
\end{wrapfigure}

\section{Stability Analysis}\label{sec:Stability analysis}
We examine whether seed-dependent data sampling affects evaluation results by replicating the AdvPCS attack under different random seed configurations. We further introduce three additional seeds to assess sensitivity to sampling variations. We use SAM3 as the victim model and evaluate on four video segmentation benchmarks: DAVIS, YouTube, MOSE, and SA-CO.
Results in \cref{fig:error_bar} summarize the variation in attack performance across seeds. The consistently small error bars indicate that AdvPCS maintains stable performance across all datasets. These results show that AdvPCS is largely insensitive to seed initialization, confirming its robustness under different random sampling conditions.

\section{Multi-point Evaluation Study}\label{sec:Multi-point Evaluation Study}
We further analyze the effect of multi-prompt evaluation settings on AdvPCS. 
Using SAM3 as the victim model, we evaluate adversarial segmentation with 1–6 input points on three datasets: YouTube, DAVIS, and MOSE. 
Results in \cref{tab:multi_point_results} show that increasing the number of prompt points provides richer spatial cues, which partially reduces the effect of adversarial perturbations.
Nevertheless, even under dense prompting, AdvPCS maintains strong attack effectiveness and consistently disrupts SAM3’s ability to produce coherent and semantically consistent segmentation. These results demonstrate that AdvPCS generalizes well across diverse prompting configurations, confirming its stability and effectiveness in multi-prompt settings.

\begin{table}[h]
\centering
\small
\setlength{\tabcolsep}{3pt}
\renewcommand{\arraystretch}{1.5}
\caption{The mIoU (\%) of AdvPCS under the multi-point evaluation mode.}
\label{tab:multi_point_results}
\begin{adjustbox}{center,width=\linewidth}
\begin{tabular}{c c || c c | c c | c c | c c}
\hline\hline
\rowcolor{gray!20}
\rule{0pt}{2.8ex} & &
\multicolumn{8}{c}{Video Segmentation} \\
\rowcolor{gray!20}
& &
\multicolumn{2}{c|}{YouTube} &
\multicolumn{2}{c|}{MOSE} &
\multicolumn{2}{c|}{DAVIS} &
\multicolumn{2}{c}{Avg} \\
\rowcolor{gray!20}
\multirow{-3}{*}{Prompt} &
\multirow{-3}{*}{Prompt Number} &
BoU & AoU & BoU & AoU & BoU & AoU & BoU & AoU \\
\hline\hline

& 1
& 85.29 & 6.29 & 69.12 & 8.17 & 69.73 & 7.89 & 74.71 & 7.45 \\
\rowcolor{gray!10}
& 2
& 82.64 & 17.61 & 71.92 & 12.67 & 82.08 & 16.84 & 78.88 & 15.71 \\
Point & 3
& 87.89 & 19.22 & 73.30 & 14.37 & 86.36 & 22.01 & 82.52 & 18.53 \\
\rowcolor{gray!10}
& 4
& 87.62 & 22.12 & 75.14 & 16.61 & 86.52 & 23.02 & 83.09 & 20.58 \\
& 5
& 88.16 & 20.83 & 75.66 & 15.44 & 86.22 & 22.85 & 83.35 & 19.71 \\

\hline
\hline
\end{tabular}
\end{adjustbox}
\vspace{-0.4cm}
\end{table}

\section{Visualization Results of Comparison Study}\label{sec:Comparison_Study_add}
\begin{figure}[!t]
    \centering
    \includegraphics[width=\linewidth]{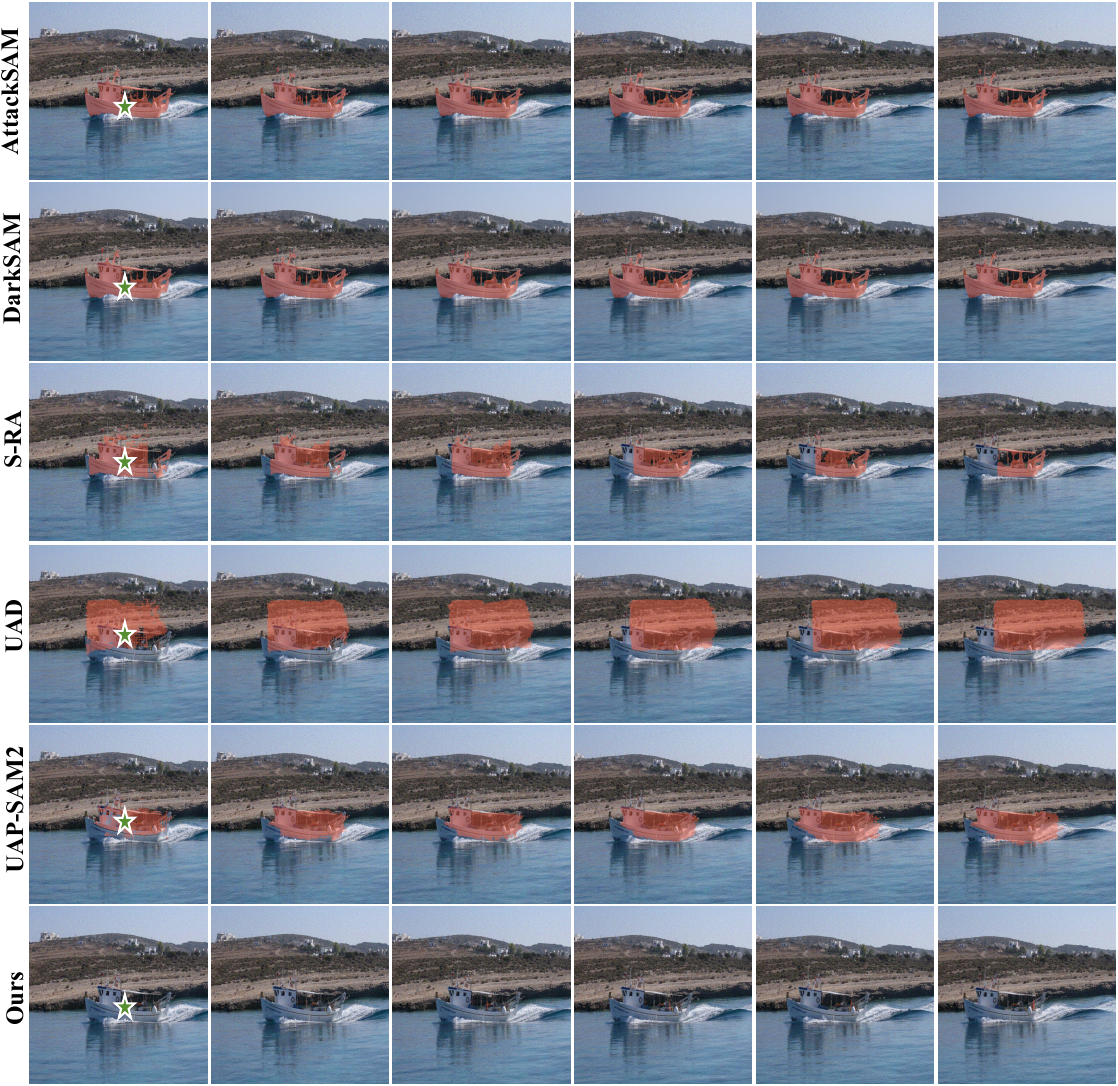}
    \caption{Visualization of adversarial examples generated by the proposed method and existing methods on the DAVIS dataset for SAM3.}
    \label{fig:attack_video_uap}
\end{figure}

We provide visualization results of comparative studies against Attack-SAM~\cite{zhang2023attack}, DarkSAM~\cite{zhou2024darksam}, S-RA~\cite{shen2024practical}, UAD~\cite{lu2024unsegment}, and UAP-SAM2~\cite{zhou2025sam2}.
We select SAM3 as the victim model and evaluate all methods on video segmentation tasks across the DAVIS dataset using point prompts.
As shown in \cref{fig:attack_video_uap}, AdvPCS outperforms all existing attack methods in video segmentation tasks.

\end{document}